\pdfoutput=1
\documentclass{article} 
\usepackage{iclr2025_conference,times}

\usepackage{amsmath,amsfonts,bm}

\def\eqref#1{equation~\ref{#1}}

\def\1{\bm{1}}

\DeclareMathAlphabet{\mathsfit}{\encodingdefault}{\sfdefault}{m}{sl}
\SetMathAlphabet{\mathsfit}{bold}{\encodingdefault}{\sfdefault}{bx}{n}

\usepackage{hyperref}
\usepackage{url}
\usepackage{amsmath}
\usepackage{bbm}
\usepackage{graphicx}
\usepackage{wrapfig}

\usepackage{tabularx}
\usepackage{booktabs}
\usepackage{multirow}
\usepackage{array}
\usepackage{makecell}
\usepackage[normalem]{ulem}
\usepackage{subcaption}
\title{Reliable and diverse evaluation of LLM medical knowledge mastery}

\author{Yuxuan Zhou, Xien Liu, Chen Ning, Xiao Zhang \& Ji Wu\\
Department of Electronic Engineering\\
Tsinghua University\\
Beijing, 100084, China \\
}

\iclrfinalcopy 

\begin{document}
\maketitle
\begin{abstract}
Mastering medical knowledge is crucial for medical-specific LLMs. However, despite the existence of medical benchmarks like MedQA, a unified framework that fully leverages existing knowledge bases to evaluate LLMs' mastery of medical knowledge is still lacking. In the study, we propose a novel framework PretexEval that dynamically generates reliable and diverse test samples to evaluate LLMs for any given medical knowledge base. We notice that test samples produced directly from knowledge bases by templates or LLMs may introduce factual errors and also lack diversity. To address these issues, we introduce a novel schema into our proposed evaluation framework that employs predicate equivalence transformations to produce a series of variants for any given medical knowledge point. Finally, these produced predicate variants are converted into textual language, resulting in a series of reliable and diverse test samples to evaluate whether LLMs fully master the given medical factual knowledge point. Here, we use our proposed framework to systematically investigate the mastery of medical factual knowledge of 12 well-known LLMs, based on two knowledge bases that are crucial for clinical diagnosis and treatment. The evaluation results illustrate that current LLMs still exhibit significant deficiencies in fully mastering medical knowledge, despite achieving considerable success on some famous public benchmarks. These new findings provide valuable insights for developing medical-specific LLMs, highlighting that current LLMs urgently need to strengthen their comprehensive and in-depth mastery of medical knowledge before being applied to real-world medical scenarios.
\end{abstract}
\section{Introduction}
Recent years have witnessed the rapid advancement of large language models (LLMs), which have exhibited potential across various domains \citep{brown2020language,ouyang2022training,touvron2023llama,openai2023gpt4,madani2023large,boiko2023autonomous}, including medicine. Solving medical problems requires LLMs to master medical factual knowledge comprehensively and in-depth. Recent studies \citep{singhal2023large,nori2023can,pal2024gemini} showed that some LLMs (e.g., GPT-4) encode medical factual knowledge, achieving outstanding performance across multiple medical benchmarks \citep{jin-etal-2019-pubmedqa,jin2021disease,medmcqa,singhal2023large,sung-etal-2021-language,meng-etal-2022-rewire}, such as MedQA. Constructed through expert annotation, these benchmarks have long been effective tools for evaluating LLMs' medical capabilities. However, they may face challenges such as becoming outdated or being possibly leaked to LLMs, which could lead to evaluations that lack reliability. Meanwhile, medical databases like UMLS \citep{bodenreider2004unified} contain extensive medical knowledge, but there is currently no unified framework that fully leverages these knowledge bases to evaluate LLMs’ mastery of medical knowledge. Therefore, we aim to bridge this gap in this study by proposing an evaluation framework that investigates LLMs' medical knowledge mastery based on any given medical knowledge base.

\begin{figure}
  \begin{minipage}{0.52\textwidth} 
    \centering
    \includegraphics[width=\linewidth]{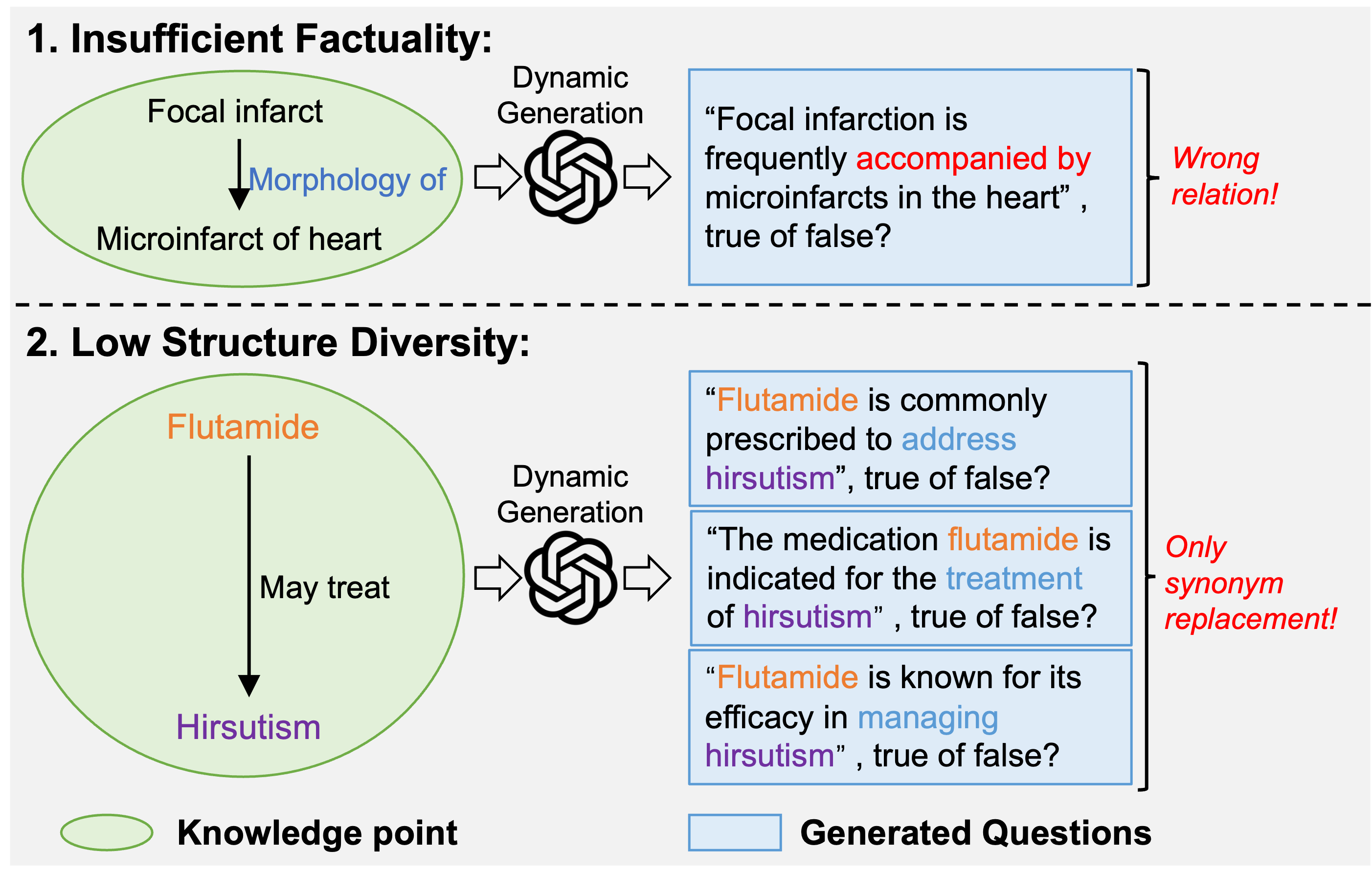} 
     \caption{Drawbacks of test samples produced directly by LLMs: (1) LLMs may introduce factual errors into generated samples; (2) Samples directly generated by LLMs exhibit low diversity.}
    \label{fig:example}
  \end{minipage}
  \hfill
  \begin{minipage}{0.46\textwidth} 
    \centering
   \includegraphics[width=\linewidth]{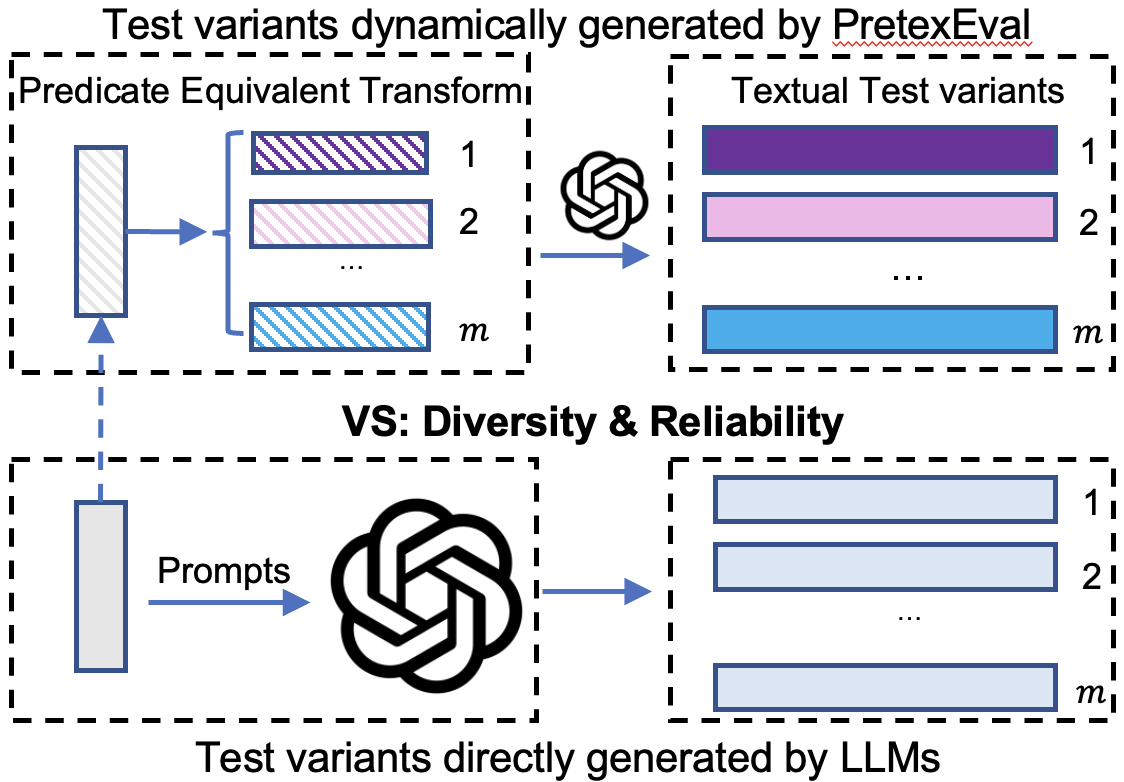}
    \caption{Schema of the proposed Predicate-to-text evaluation method (Top) compared with directly generating test variants by LLMs (Bottom).}
    \label{fig:principle}
  \end{minipage}
\end{figure}

Evaluating LLMs using medical knowledge bases requires generating textual test samples from structured knowledge. A straightforward method is to prompt LLMs to directly generate test samples based on specific knowledge points. However, this method has two drawbacks as illustrated in Figure \ref{fig:example}: (1) \textbf{insufficient factuality}: factual errors (e.g., incorrect relations) may be introduced during LLM generation process, affecting the reliability of evaluation; and (2) \textbf{low structure diversity}: samples generated from the same knowledge point primarily differ in wording (e.g., synonym replacement) rather than in expression structure, compromising the diversity of evaluation. 

The purpose of this study is to develop a unified evaluation framework that dynamically generates reliable and diverse test samples from medical knowledge bases to probe LLMs’ mastery of medical factual knowledge. Given that medical factual knowledge primarily involves relationships between medical entities, it can be effectively expressed through predicates.
Inspired by this, we propose a \textbf{Pre}dicate-to-\textbf{tex}t \textbf{Eval}uation method (\textbf{PretexEval}) that dynamically generates reliable and structurally diverse test samples based on the medical knowledge points from knowledge bases. Figure \ref{fig:principle} presents the schema of our method. Specifically, we first express each knowledge point using a predicate expression. Then, we derive a series of predicate variants from this predicate expression through predicate equivalent/implication transformation\footnote{For the sake of convenience, we refer to both equivalent and implication transformations collectively as predicate equivalent transformations in the following sections, without making a distinction between them.}. Such predicate transformation process enhances the structural diversity of generated test samples and also effectively prevents the introduction of factual errors. Finally, we use a prototype-based method to convert these variants back into the textual space to create the test samples. This method first transforms the predicate variants into prototype samples using templates to ensure reliability, and then rephrases these prototypes with LLMs to enhance the readability and lexical diversity of the generated samples. It is worth noting that the proposed evaluation method is highly \textbf{versatile} and can be applied to any medical knowledge base with minimal adjustments to evaluate LLMs’ mastery of the knowledge it contains.

In our study, we employ the proposed evaluation framework to systematically investigate the mastery of medical knowledge among 12 well-known LLMs, using two medical knowledge bases that contain essential information for clinical diagnosis and treatment.
Experimental results indicate that the performance of current LLMs on test samples generated by our method is significantly lower than on samples directly generated by handcrafted templates or by prompting an LLM. Furthermore, these evaluated LLMs exhibit notable inconsistency in handling test samples derived from the same knowledge point. These findings indicate that, despite their impressive performance on several medical benchmarks, current LLMs have not fully mastered the medical knowledge essential for real-world clinical tasks. Therefore, they may require additional training before being applied in real-world medical scenarios to further enhance their comprehensive and in-depth mastery of medical knowledge. The codes and datasets will be released to facilitate future study. Our contributions are summarized as follows:
\begin{itemize}
    \item We introduce PretexEval, a predicate-to-text method that dynamically generates reliable and structurally diverse test samples based on any given medical knowledge base. 
    \item Using the proposed method, we systematically investigate the medical factual knowledge mastery of 12 well-known LLMs based on two knowledge bases closely related to clinical diagnosis and treatment. 
    \item Our findings reveal that current LLMs have not yet comprehensively and deeply mastered medical knowledge, underscoring the urgent need to improve their medical knowledge mastery before applying them to real-world medical tasks.
\end{itemize}
\section{Related Work}
\paragraph{LLM Medical Evaluation}
Current medical evaluation benchmarks for LLMs can be divided into two categories: (1) QA datasets that evaluate LLMs' comprehensive medical capabilities with questions collected from medical literature \citep{jin-etal-2019-pubmedqa}, exams \citep{jin2021disease,medmcqa}, or online websites \citep{singhal2023large}; (2) datasets for probing LLM medical knowledge mastery \citep{sung-etal-2021-language,meng-etal-2022-rewire}. These static benchmarks are meticulously created by medical experts and possess high reliability. However, they may face problems such as becoming outdated or leaked to LLMs, affecting the comprehensiveness of evaluation. While constructing new benchmarks can alleviate these problems, they will also become obsolete over time. 
\paragraph{Dynamic Evaluation Schema}
Several studies have proposed dynamic evaluation methods that automatically generate new test samples, effectively avoiding data obsolescence and leakage issues. Some works leverage algorithms to dynamically generate test samples for specific tasks, such as mathematics \citep{zhu2024dyval} and SQL execution \citep{lei2023s3eval}. Others \citep{zhu2023clean,zhu2024dyval2} generate test samples by paraphrasing existing benchmarks. However, there is currently no related work utilizing dynamic evaluation methods to evaluate LLMs based on knowledge bases. To our knowledge, our proposed method is the first to apply the dynamic evaluation schema for evaluating LLMs' mastery of factual knowledge based on knowledge bases. 
\section{Method}
\subsection{Evaluation Schema}
In this section, we introduce the schema of our PretexEval method, which generates structural diverse and reliable test samples for LLM factual knowledge evaluation. Given a knowledge point ${\mathrm{P}}$, a straightforward idea is to directly generate a test sample using an LLM:
\begin{equation}
    \mathrm{S} = \operatorname{G}_{\text{LLM}}(\mathrm{P})
\end{equation}
Here, $\operatorname{G}_{\text{LLM}}$ denotes the LLM generation process, and $\mathrm{S}$ refers to the generated test sample. As introduced above, $\operatorname{G}_{\text{LLM}}$ may create samples that lack diversity and reliability. In contrast, our method first expresses the knowledge point using a predicate expression and then derives a series of variants via predicate equivalent transformation:
\begin{gather}
    \mathrm{q} = \operatorname{T}_{\text{text2pre}}(\mathrm{P})\\
    [\mathrm{v}_1,\mathrm{v}_2,\cdots,\mathrm{v}_m] = \operatorname{T}_{\text{Eq}}(\mathrm{q})
\end{gather}
Here, $\operatorname{T}_{text2pre}$ denotes a mapping that projects the original knowledge point $P$ into the predicate expression $\mathrm{q}$. $\operatorname{T}_{\text{Eq}}$ refers to the predicate equivalent transformation, and $\{\mathrm{v}_i\}_{i=1}^{m}$ are the variants derived from the original expression $\mathrm{q}$. The property of predicate equivalent transformation ensures the reliability of these variants, provided that the original expression $\mathrm{q}$ is true:
\begin{equation}
    (\mathrm{q}=\text{True}) \Rightarrow (\mathrm{v}_i=\text{True}),\quad 1\leq i \leq m
\end{equation}
Finally, we convert each predicate variant back to a textual test sample for evaluation:
\begin{equation}
    \mathrm{S}_i = \operatorname{T}_{\text{pre2text}}(\mathrm{v}_i),\quad 1\leq i \leq m
\end{equation}
Here, $\operatorname{T}_{pre2text}$ maps each predicate variant $\mathrm{v}_i$ into a corresponding test sample (textual variant). Since these samples are derived from predicate variants with diverse structures, the predicate-text duality ensures they exhibit substantial diversity while maintaining reliability.
\begin{figure}[t]
    \centering
    \includegraphics[width=0.97\textwidth]{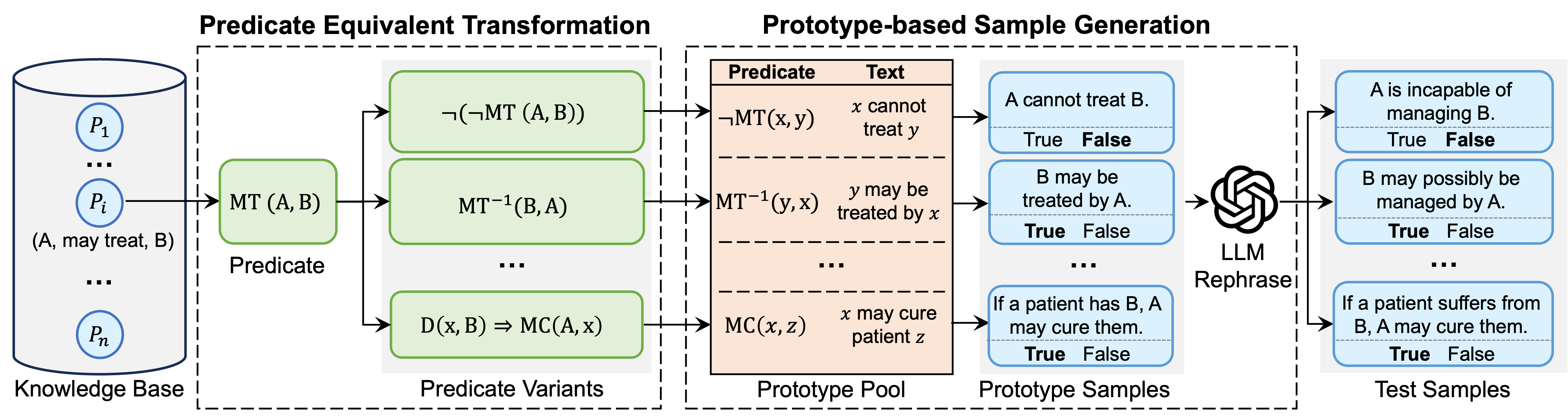}
    \caption{An overview of the proposed PretexEval framework, which dynamically generates test samples from any medical knowledge base for evaluating LLMs’ medical knowledge mastery.}
    \label{fig:overview}
\end{figure}
\subsection{Evaluation Framework}\label{sec:framework}
Building on the proposed evaluation schema, we develop a novel evaluation framework to evaluate LLMs' mastery of medical factual knowledge comprehensively. Figure \ref{fig:overview} presents an overview of this framework.
\subsubsection{Predicate Variant Generation}
A single knowledge point (i.e., knowledge triplet in knowledge bases) can be denoted as $\mathrm{P}=(\mathrm{h}, \mathrm{r}, \mathrm{t})$, where $\mathrm{h}$, $\mathrm{r}$, and $\mathrm{t}$ refer to the head entity, the relation, and the tail entity, respectively. In predicate logic, such a relation can be effectively presented by:
\begin{equation}
    \mathrm{q}=\mathcal{R}(\mathrm{h},\mathrm{t})
\end{equation}

Here, $\mathcal{R}(\mathrm{x}, \mathrm{y})$ is a predicate derived from the relation $r$, representing the statement "$\mathrm{x}$ has the relation $\mathrm{r}$ with $\mathrm{y}$", where $\mathrm{q}$ represents its value at the point $(\mathrm{h}, \mathrm{t})$. Next, the framework employs three types of equivalent transformations that are widely employed in practical medical applications, including:
\begin{itemize}
    \item \textbf{Inversion} ($\mathcal{R}^{-1}(\mathrm{t},\mathrm{h}))$): The inverse expression presents the original expression from another direction. For example, if the statement ``Drug A may treat disease B" holds, then ``Disease B's prescribed drug includes drug A" also holds.
    \item \textbf{Instantiation} ($\mathcal{P}(\mathrm{h},\mathrm{x})\Rightarrow\mathcal{Q}(\mathrm{x},\mathrm{t})$): This type of transformation applies a general knowledge point to a specific case. For example, the statement ``Drug A may treat disease B" can be instantiated as ``If a patient has disease B, drug A may cure them." Such transformation is commonly used in disease diagnosis and treatment.
    \item \textbf{Double Negation} ($\neg(\neg \mathcal{R}(\mathrm{h},\mathrm{t}))$): The double negation rule is widely utilized to obtain logically equivalent expressions. In our framework, this rule is applied to construct \textbf{negative} expressions. For example, if ``Drug A may treat disease B" holds, then ``Drug A cannot treat disease B" must be false.
\end{itemize}
It is noteworthy that these three types of transformation can be further combined to produce additional expressions based on the transitive property of predicate equivalent transformation. As a result, a total of $m$ variants are generated in this process:
\begin{equation}
    \mathrm{v}_i=\operatorname{T}^{i}_{\text{Eq}}(\mathcal{R}(\mathrm{h},\mathrm{t})),\quad 1\leq i \leq m
\end{equation}
where $\operatorname{T}^{i}_{\text{Eq}}$ denotes the $i^{th}$ predicate equivalent transformation.

\subsubsection{Textual Sample Generation}
Finally, the predicate variants are converted back into textual samples for LLM evaluation. A straightforward method is template rephrasing; however, the generated samples may lack fluency, which may affect LLM’s understanding. Another approach is to prompt LLMs to generate test samples directly from predicates. However, considering that LLMs may not fully understand predicate forms, this method can also introduce factual errors. To address this issue, we designed a prototype-based sample generation method. Specifically, for each predicate variant $\operatorname{T}^{i}_{\text{Eq}}(\mathcal{R}(\mathrm{h},\mathrm{t}))$, we initially retrieve the corresponding prototype from a pre-constructed prototype pool based on the predicate $\operatorname{T}^{i}_{\text{Eq}}\cdot\mathcal{R}$. For predicate variants obtained through double negation, we retrieve prototypes based on their negated form (i.e., single negation form) to generate \textbf{negated samples} for LLM evaluation. Subsequently, the prototype is instantiated by the arguments $(\mathrm{h},\mathrm{t})$. The instantiated prototype precisely conveys the predicate variant in the textual space. Finally, the prototype is further rephrased by an LLM to obtain the final test sample $\mathrm{S}_i$. Since current LLMs possess strong language capabilities and seldom make mistakes in sentence rephrasing, the proposed sample generation method can ensure the reliability and diversity of the generated samples.

\subsubsection{Evaluation Metrics}\label{sec:eval}
In our framework, we evaluate LLMs using statement verification tasks, asking them to determine whether a given statement is true or false:
\begin{equation}
    \operatorname{Score}(\mathrm{M},\mathrm{S}_i)= \mathbbm{1}(\mathrm{M}(\mathrm{S}_i)=l_i), 1\leq i \leq m
\end{equation}
Here, $\mathrm{M}$ is the evaluated LLM, $\mathrm{S}_i$ is the textual variant (statement) generated by our framework, and $\mathrm{M}(\mathrm{S}_i) \in {\{\textbf{T}, \textbf{F}\}}$ denotes LLM's prediction for $\mathrm{S}_i$. $l_i \in {\{\textbf{T}, \textbf{F}\}}$ is the label of $\mathrm{S}_i$, and the function $\mathbbm{1}(\cdot)$ is a characteristic function that equals 1 when the enclosed expression is true, and 0 otherwise.

For a dataset with $n$ knowledge points $\{\mathrm{P}_j\}_{j=1}^{n}$, we initially use the metric \textit{average accuracy} to compute the accuracy across all test samples:
\begin{equation}
    a_{\text{avg}} = \frac{1}{n}\frac{1}{m}\sum_{j=1}^{n}\sum_{i=1}^{m}\operatorname{Score}(\mathrm{M},\mathrm{S}^{j}_i)
\end{equation}
Here, $\mathrm{S}^{j}_i$ denotes the $i^{th}$ test sample derived from the $j^{th}$ knowledge point $\mathrm{P}_j$. While this metric is widely applied in various benchmarks, it cannot evaluate the \textbf{consistency} of LLMs in predicting all test samples derived from the same knowledge point, which is crucial for high-risk applications in the medical domain. Therefore, we also utilize another metric, \textit{joint accuracy}, which considers a knowledge point as mastered if \textbf{all the related samples} are predicted correctly:
\begin{equation}
   a_{\text{joint}} = \frac{1}{n}\sum_{j=1}^{n}\prod_{i=1}^{m}\operatorname{Score}(\mathrm{M},\mathrm{S}^{j}_i)
\end{equation}
By applying these metrics, we can achieve a comprehensive evaluation of LLMs' mastery of medical factual knowledge.

\section{Experiments}
\subsection{Experiment Setup}\label{sec:setup} 
\paragraph{Datasets Introduction}
To validate the effectiveness of our proposed framework PretexEval, we conduct a systematic evaluation on LLMs' medical knowledge mastery with PretexEval using two knowledge bases: a biomedical knowledge base MedLAMA \citep{meng-etal-2022-rewire} and a clinical knowledge base DiseK \citep{zhou2024multifaceteval}. MedLAMA is a large-scale biomedical knowledge base consisting of 39,053 knowledge triplets that encompass 19 relations among medical entities such as diseases, genes, cells, and tissues, all meticulously selected from the UMLS Metathesaurus \citep{bodenreider2004unified} to ensure high quality. DiseK is a clinical knowledge base that contains 24,413 triplets, covering 1,000 high-frequency diseases across four key relations related to disease diagnosis and treatment. Mastering the knowledge contained within these databases is essential for LLMs to be effectively applied in real medical scenarios. It is important to highlight that our framework can also be applied to other medical knowledge bases with minimal adjustments to evaluate medical knowledge of other types (e.g., drug-related knowledge); we consider this as future work.

Considering computational costs and dataset size, we select a subset from each dataset for evaluation. Specifically, we randomly select a single entity from the corresponding tail entities for each pair of a head entity and a relation. This approach aims to reduce the evaluation scale while maximizing the diversity of the evaluated knowledge. We also excluded two relations in MedLAMA, which are the inversion of the other two relations in MedLAMA. Furthermore, for each head-relation pair $(\mathrm{h},\mathrm{r})$, we randomly sample a negative entity $\mathrm{c}$ that satisfies $\neg \mathcal{R}(\mathrm{h},\mathrm{c})$ to create a negative triplet $(\mathrm{h},\mathrm{r},\mathrm{c})$. Test samples generated from this negative triplet possess a similar structure to those generated from the positive triplet but with opposite labels. By introducing negative triplets, we can further evaluate the ability of LLMs to discern non-knowledge, which is also essential for practical application. Appendix \ref{apdix:dataset} provides detailed statistics about these knowledge bases and relation types within (Table \ref{tab:relmedlama}, \ref{tab:reldisek}).
\paragraph{Method Setting}
To ensure the diversity of evaluation, we combined the three types of predicate transformation and generated $m=8$ expressions (variants) for each knowledge point, including the original expression. We crafted a prototype for each combination of relation and predicate transformation type to generate test samples. Moreover, we utilize Llama3-70B-Instruct \citep{llama3modelcard} to rephrase the instantiated prototypes because of its strong performance. We have also tried other rephrasing LLMs and obtained similar evaluation results (see Appendix \ref{apdix:reph}). More details of the predicate transformation, prototypes, and the prompt format are provided in Appendix \ref{apdix:framework}.

For LLM evaluation, we employ the popular 5-shot in-context learning setting \citep{brown2020language}, where five examples are presented before the test sample, guiding LLMs to produce answers in consistent format with the provided examples. We calculate the average and joint accuracies (introduced in Sec \ref{sec:eval}) for each LLM. Appendix \ref{apdix:eval} provides more details, including the prompt format.
\paragraph{Baselines}
We initially compare our method with the method that directly generates test samples by paraphrasing the knowledge with templates (denoted as Direct). We also implemented a dynamic evaluation baseline (named as \textbf{LLMEval}) that directly generates test samples from triplets using an LLM. Specifically, we prompt Llama3-70B-Instruct\footnote{We choose the same LLM utilized in our framework to make a fair comparison.} to generate $m=8$ statements, presenting the given triplet in different ways. We carefully crafted the prompt to ensure maximum diversity in generated samples. Appendix \ref{apdix:baseline} details the prompt and other settings.
\paragraph{Evaluated LLMs}
In our study, we evaluate 12 well-known general and medical-specific LLMs: (1) general LLMs: Gemma-7B \citep{team2024gemma}, Llama2 (7B,70B) \citep{touvron2023llama}, Llama3 (8B,70B) \citep{llama3modelcard}, Vicuna (7B,13B) \citep{zheng2023judging}, GPT-3.5-turbo \citep{ouyang2022training}, and the latest GPT-4o \citep{gpt-4o}; (2) medical-specific LLMs: ClinicalCamel-70B \citep{toma2023clinical}, Meditron-70B \citep{chen2023meditron} and Med42-70B \citep{med42}. For cost considerations, we evaluate GPT-4 on a sampled subset containing 200 knowledge triplets for each dataset.
\subsection{Results}
\subsubsection{Comparison Study}
\begin{table}[t]
\centering
\setlength{\tabcolsep}{2mm}{
\begin{tabular}{l|ccc|ccc}
\hline
\multirow{2}{*}{Model} & \multicolumn{3}{c|}{MedLAMA}                                                              & \multicolumn{3}{c}{DiseK}                                                                 \\
                       & \multicolumn{1}{c}{Direct} & \multicolumn{1}{c}{LLMEval} & \multicolumn{1}{c|}{PretexEval}   & \multicolumn{1}{c}{Direct} & \multicolumn{1}{c}{LLMEval} & \multicolumn{1}{c}{PretexEval}    \\ \hline

Llama2-7B              & +$6.4$  & +$8.3\textsubscript{↑29.7\%}$  & +$\underline{3.1}\textsubscript{↓52.3\%}$  & +$11.7$ & +$2.7\textsubscript{↓76.6\%}$  & +$\underline{2.8}\textsubscript{↓76.3\%}$  \\
Vicuna-7B              & +$26.4$ & +$18.0\textsubscript{↓31.7\%}$ & +$\underline{7.5}\textsubscript{↓71.5\%}$  & +$9.9$  & +$10.9\textsubscript{↑9.7\%}$  & +$\underline{3.9}\textsubscript{↓60.5\%}$  \\
Vicuna-13B             & +$27.0$ & +$19.3\textsubscript{↓28.5\%}$ & +$\underline{10.7}\textsubscript{↓60.5\%}$ & +$12.5$ & +$7.4\textsubscript{↓40.2\%}$  & +$\underline{5.7}\textsubscript{↓53.9\%}$  \\
Gemma-7B               & +$23.3$ & +$11.1\textsubscript{↓52.3\%}$ & +$\underline{9.4}\textsubscript{↓59.5\%}$  & +$9.0$  & +$\underline{4.8}\textsubscript{↓46.5\%}$  & +$5.0\textsubscript{↓45.0\%}$  \\
Llama3-8B              & +$28.5$ & +$19.1\textsubscript{↓33.1\%}$ & +$\underline{16.6}\textsubscript{↓41.8\%}$ & +$17.9$ & +$15.3\textsubscript{↓14.3\%}$ & +$\underline{9.3}\textsubscript{↓48.3\%}$  \\ \hline
Llama2-70B             & +$32.0$ & +$19.2\textsubscript{↓39.9\%}$ & +$\underline{13.8}\textsubscript{↓56.9\%}$ & +$20.5$ & +$17.3\textsubscript{↓15.7\%}$ & +$\underline{9.0}\textsubscript{↓56.0\%}$  \\
ClinicalCamel-70B      & +$34.8$ & +$23.7\textsubscript{↓31.9\%}$ & +$\underline{21.9}\textsubscript{↓37.2\%}$ & +$24.5$ & +$20.6\textsubscript{↓15.7\%}$ & +$\underline{16.1}\textsubscript{↓34.4\%}$ \\
Meditron-70B           & +$29.4$ & +$20.0\textsubscript{↓32.1\%}$ & +$\underline{14.7}\textsubscript{↓49.8\%}$ & +$21.1$ & +$12.8\textsubscript{↓39.4\%}$ & +$\underline{10.2}\textsubscript{↓51.5\%}$ \\
Med42-70B              & +$31.8$ & +$\underline{19.3}\textsubscript{↓39.3\%}$ & +$20.0\textsubscript{↓37.1\%}$ & +$23.3$ & +$19.1\textsubscript{↓18.1\%}$ & +$\underline{14.8}\textsubscript{↓36.6\%}$ \\
Llama3-70B             & +$\textbf{36.6}$ & +$26.9\textsubscript{↓26.5\%}$ & +$\underline{26.9}\textsubscript{↓26.6\%}$ & +$29.7$ & +$28.2\textsubscript{↓5.1\%}$  & +$\underline{20.9}\textsubscript{↓29.7\%}$ \\\hline
GPT-3.5-turbo          & +$32.1$ & +$26.7\textsubscript{↓16.9\%}$ & +$\underline{16.2}\textsubscript{↓49.7\%}$ & +$23.5$ & +$17.6\textsubscript{↓25.4\%}$ & +$\underline{10.3}\textsubscript{↓56.4\%}$ \\
GPT-4o$^*$             & +$35.8$ & +$\textbf{34.0}\textsubscript{↓4.9\%}$  & +$\underline{\textbf{31.7}}\textsubscript{↓11.5\%}$ & +$\textbf{31.3}$ & +$\textbf{29.9}\textsubscript{↓4.3\%}$  & +$\underline{\textbf{26.7}}\textsubscript{↓14.5\%}$ \\ \hline
\end{tabular}}
\caption{Performance (\textbf{average accuracy}) of LLMs evaluated on datasets directly generated by template paraphrasing (Direct), datasets directly generated by LLM (LLMEval), and datasets generated by \textbf{our framework (PretexEval)}. We report the gain relative to random guessing (50\%) and the relative performance degradation compared to the Direct results. Bold: Best performance under the same evaluation method; Underline: LLM achieved the lowest performance on this evaluation method. $*$GPT-4 is evaluated on sampled subsets for cost considerations.}
\label{tab:performance}
\end{table}
We first conduct a comparison study across different evaluation methods and LLMs. Table \ref{tab:performance} lists LLMs' performance (average accuracy) on the MedLAMA and DiseK knowledge bases evaluated by different methods. We also conduct a fine-grained analysis on LLMs performance across knowledge types, which is provided in Appendix \ref{apdix:types} due to the space limit. The results demonstrate that all evaluated LLMs achieve much lower performance on datasets generated by PretexEval compared to the original datasets. This suggests that \textbf{dynamically generating multiple samples for each knowledge point can significantly enhance the comprehensiveness of evaluation}. Moreover, compared to datasets directly generated by an LLM (LLMEval), almost all LLMs achieve lower performance on datasets created by PretexEval, with some models (e.g., Llama2-7B and Llama2-70B) experiencing over 50\% degradation. These findings suggest that \textbf{PretexEval is capable of generating test samples that are more diverse than those directly generated by LLMs.}

Among all the evaluated LLMs, GPT-4o outperforms the others across almost all datasets and evaluation methods, achieving performance gains (relative to random guessing (50\%)) of 31.7 and 26.7 evaluated by PretexEval. On open-sourced LLMs, Llama3-70B and Llama3-8B performs best on PretexEval-generated datasets compared to LLMs with similar parameter scales. It is worth noting that Llama3-8B even slightly surpassing the 10x larger Llama2-70B. These results indicate that \textbf{Llama3 model series encodes significantly more medical knowledge than other evaluated LLMs}. Additionally, while some medical-specific LLMs (ClinicalCamel, Med42) perform similarly to their backbone model (Llama2-70B) on original datasets, they notably outperform the latter by around 7\% on PretexEval-generated datasets. This suggests that \textbf{training on medical corpora can notably improve the depth of medical knowledge mastery}.

\begin{figure}[t]
    \centering
    \includegraphics[width=\textwidth]{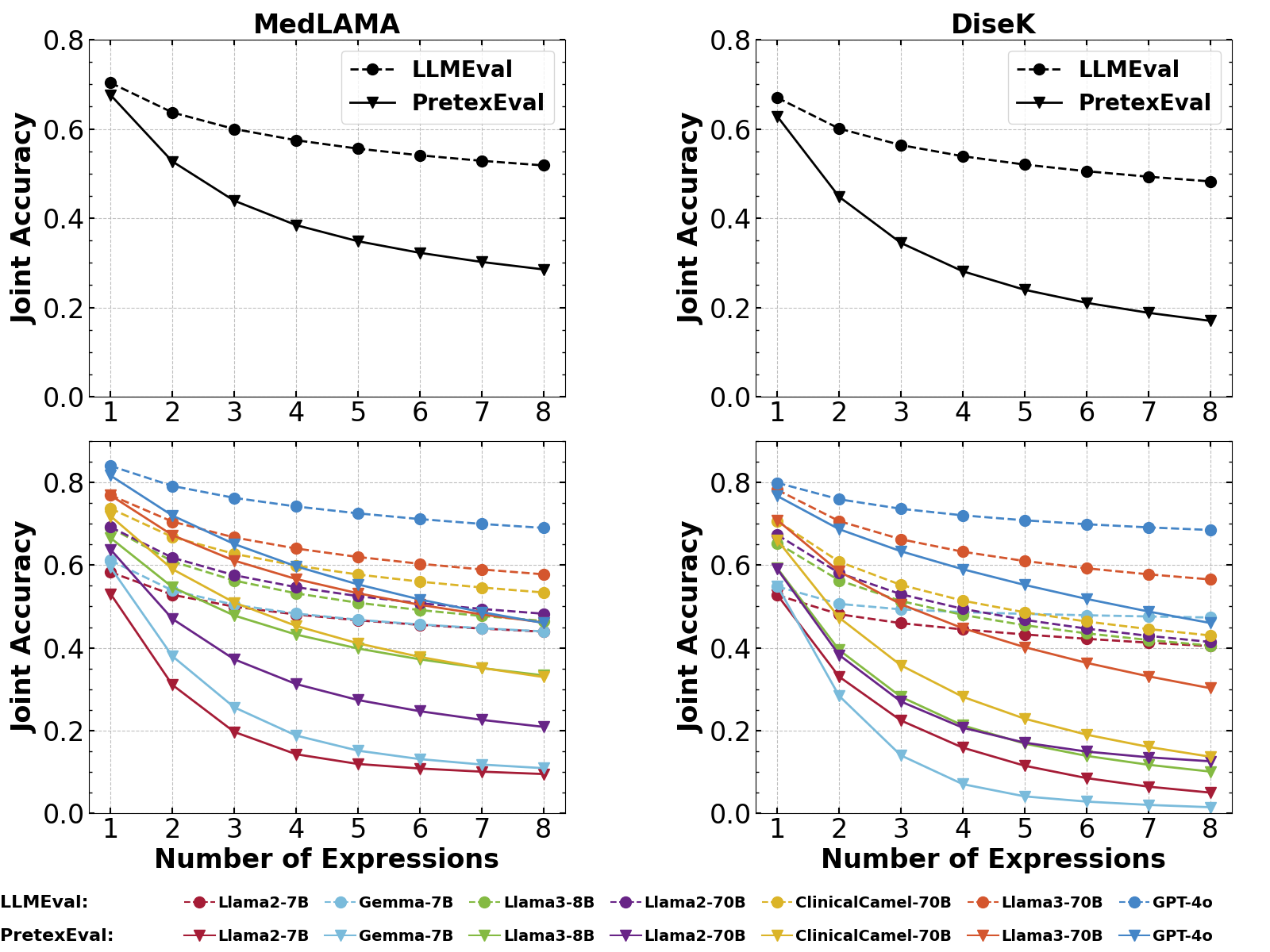}
    \caption{Performance (\textbf{joint accuracy}) of 7 typical LLMs evaluated by increasing the number of expressions per knowledge point. Top: overall performance trend averaged across LLMs; bottom: detailed performance for each LLM. To eliminate the impact of sample addition orders, we enumerate all possible orders and averaged the results, where the value at $x=i$ corresponds to the expected joint accuracy evaluated with any $i$ samples.}
    \label{fig: joint}
\end{figure}
We also study the joint accuracies of LLMs evaluated by increasing numbers of expressions per knowledge point. The results of seven typical LLMs are illustrated in Figure \ref{fig: joint}, with the full results provided in Appendix \ref{apdix:joint}. We observe that the results from LLMEval and PretexEval are quite close when using a single sample for evaluation. However, as the number of test samples increases, the difference between the results from the two methods grows notably larger. This phenomenon indicates that \textbf{current LLMs generally exhibit significant lower consistency when confronted with structurally diverse test samples generated by our method} compared to samples directly generated by LLMs. Moreover, as the number of expressions increases, GPT-4o and Llama3-70B exhibits a slower decline in performance compared to other LLMs, indicating a more consistent understanding of diverse expression structures from the same knowledge points. Nevertheless, there is still room for improvement in current LLMs' mastery of medical knowledge.
\subsubsection{Effectiveness Analysis}
\paragraph{Effect of framework components}
First, we conduct an ablation study to analyze the contribution of each component in our proposed framework. Table \ref{tab:ablation} presents the ablation results of three typical LLMs, and the full results are listed in Appendix \ref{apdix:ablation}. Here, we focus on the predicate equivalent transformation and the LLM rephrasing process in the prototype-based generation module that are designed to increase the diversity of test samples. We observe that removing these two modules results in higher evaluation performance, especially when the predicate equivalent transformation module was removed (around 7\% on Llama3-70B). These results indicate that the \textbf{predicate equivalent transformation contributes most to the evaluation diversity in the proposed framework}.
\paragraph{Effect of Predicate Transformation Types}
\begin{table}[t]
    \centering
    \setlength{\tabcolsep}{1mm}{
\begin{tabular}{llccc}
\hline
Knowledge Base & Method                & ClinCamel-70B & Llama3-70B & GPT-4o \\ \hline
\multirow{3}{*}{MedLAMA}        & PretexEval (Ours)                        & +21.9             & +26.9      & +31.7  \\
                                & w/o Predicate Transformation & +30.6             & +33.0      & +36.0  \\
                                & w/o LLM Rephrasing                       & +22.8             & +30.4      & +33.8  \\ \hline
\multirow{3}{*}{DiseK}          & PretexEval (Ours)                        & +16.1             & +20.9      & +26.7  \\
                                & w/o Predicate Transformation & +23.1             & +27.8      & +29.3  \\
                                & w/o LLM Rephrasing                       & +18.0             & +24.0      & +30.4  \\ \hline
\end{tabular}}
    \caption{Ablation results of three typical LLMs for key components of the proposed PretexEval framework. Predicate Transformation: the predicate equivalent transformation module; LLM Rephrasing: the LLM rephrasing module in the prototype-based generation module. }
    \label{tab:ablation}
\end{table}
\begin{table}[t]
    \centering
    \setlength{\tabcolsep}{1mm}{
    \begin{tabular}{llccc}
\hline
Knowledge Base & Transformation Type  & ClinCamel-70B & Llama3-70B & GPT-4o \\ \hline
\multirow{4}{*}{MedLAMA}        & Direct                               & +30.6             & +33.0      & +36.0  \\
                                & +Inversion                           & +30.3             & +31.8      & +34.3  \\
                                & +Inversion+Double Negation           & +23.2             & +28.6      & +33.6  \\
                                & +All                                 & +14.7             & +26.9      & +31.7  \\ \hline
\multirow{4}{*}{DiseK}          & Direct                               & +23.1             & +27.8      & +29.3  \\
                                & +Inversion                           & +22.4             & +27.5      & +29.8  \\
                                & +Inversion+Double Negation           & +17.9             & +22.3      & +26.8  \\
                                & +All                                 & +16.1             & +20.9      & +26.7  \\ \hline
\end{tabular}}
    \caption{Ablation results of three typical LLMs for different predicate transformations in PretexEval. Each row represents a cumulative experiment, adding one transformation type at a time, with “All” indicating the combination of instantiation, inversion, and double negation.}
    \label{tab:types}

\end{table}
We further conduct a fine-grained analysis of the predicate transformation types applied in our framework, with results presented in Table \ref{tab:types}. Experimental results show that LLM performance continually declines as more predicate transformation types are added, indicating their effectiveness. Furthermore, the inclusion of double negation (+DN) leads to a more significant performance degradation (around 5\%) than other implication types. This suggests that current LLMs exhibit relatively \textbf{less proficiency in understanding negated expressions} compared to instantiated and inverted statements of medical knowledge.
\begin{figure}[t]
    \centering
    \includegraphics[width=\textwidth]{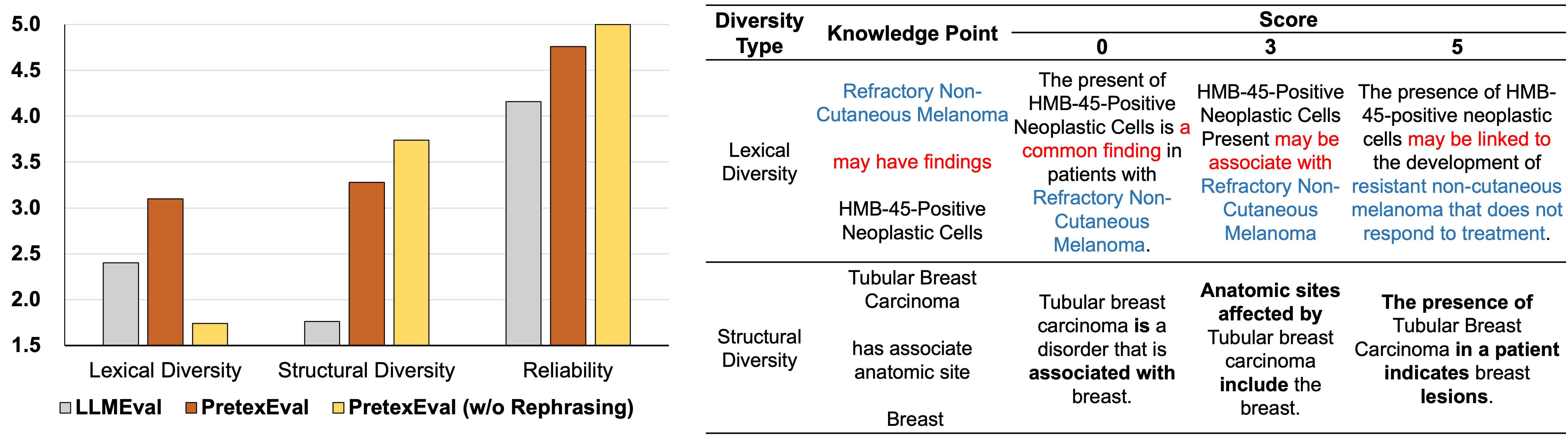}
    \caption{Left: Results of the human analysis on the reliability and diversity (lexical, structural) of samples generated by different methods; Right: Text examples in different grades of diversity.}
    \label{fig: human}
\end{figure}
\paragraph{Reliability \& Diversity of Generated Samples}
We further conduct a human analysis to investigate the reliability and diversity of samples generated by different methods. Specifically, we randomly sample 50 knowledge triplets from MedLAMA and have two experienced doctors to score the test samples regarding their lexical diversity, structural diversity, and reliability by comparing with the original knowledge triplet. The analysis results and examples in different grades are illustrated in Figure \ref{fig: human}, with more details (scoring criteria) of this analysis provided in Appendix \ref{apdix:quality}. We observe that, before the rephrasing process, the prototype samples generated by PretexEval exhibit high structural diversity and reliability but have lower lexical diversity compared to other methods. Although the samples generated by LLMEval achieve relatively high lexical diversity, they significantly lack structural diversity and are also less reliable than the prototype samples. After rephrasing, the PretexEval-generated samples maintain high structural diversity and reliability, while also achieving much higher lexical diversity. Our findings indicate that the proposed PretexEval method is capable of generating reliable and diverse test samples based on knowledge bases.
\begin{figure}[h]
    \centering
    \includegraphics[width=\linewidth]{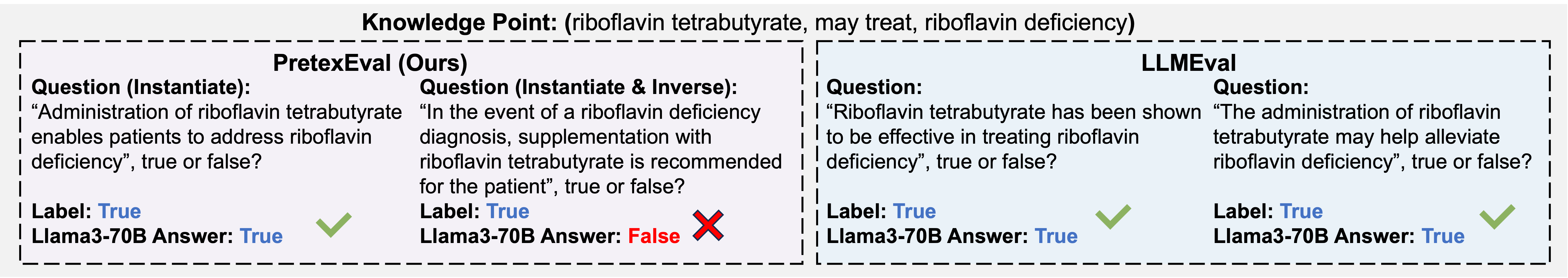}
    \caption{A case of evaluating LLMs using PretexEval compared with the LLMEval method.}
    \label{fig: case}
\end{figure}
\subsubsection{Case Study}
We also conduct a case study on our evaluation. Figure \ref{fig: case} illustrates the text samples generated by PretexEval in comparison with those generated by the LLMEval method, along with the LLMs’ responses. The case shows that the proposed PretexEval framework generates test samples that are much diverse than those directly generated by LLMs, enabling a more comprehensive evaluation.

\section{Conclusion and Discussion}
In this paper, we propose PretexEval, a novel evaluation framework that dynamically generates reliable and diverse test samples based on medical knowledge bases to probe LLMs' mastery of medical factual knowledge. The proposed framework is highly generalizable and can be applied to any medical knowledge base with minor adjustments. We validated the effectiveness of PretexEval by conducted a systematic evaluation based on two medical knowledge bases. The experimental results show that the performance of current LLMs evaluated by the proposed framework is much worse than their performance on public medical benchmarks. Furthermore, these LLMs exhibit inconsistency in understanding diverse expressions derived from the same medical knowledge point. These findings suggest that current LLMs have not fully mastered medical knowledge, which may be one of the potential reasons for their unsatisfactory performance on real-world medical scenarios. Our study may also shed light on developing medical foundation models. For example, incorporating content that presents the same medical knowledge in diverse ways into the training data may improve LLMs' consistency and comprehensiveness in understanding medical concepts. In the future, we aim to integrate this framework with other evaluation forms (e.g., question answering) and medical knowledge bases to conduct a more comprehensive evaluation of LLM medical knowledge mastery.

\bibliography{iclr2025_conference}
\bibliographystyle{iclr2025_conference}

\appendix
\section{Details of Datasets}\label{apdix:dataset}
We validate the proposed framework on two datasets: a biomedical evaluation benchmark, MedLAMA, and a disease-centric clinical knowledge base, DiseK. Given the large scale of these datasets, we sample a subset of knowledge points from each by selecting a single tail entity for each 1-to-N relation. Additionally, we sample negative triplets to increase the evaluation difficulty. Table \ref{tab:relmedlama} and \ref{tab:reldisek} list the relation types involved in the sampled datasets. The sampled MedLAMA dataset includes 1,000 positive triplets and 1,000 negative triplets for each relation, while the detailed statistics for DiseK are presented in Table \ref{tab:statistics}.
\begin{table}[h]
\centering
\begin{tabular}{m{4cm}m{9cm}}
\hline
Relation Type                                 & Description                                                                                               \\ \hline
associated morphology of                      & A particular morphology (structural feature or form) is associated with another concept, often a disease. \\
disease has abnormal cell                     & A disease is characterized by the presence of abnormal cells.                                             \\
disease has associated anatomic site          & A disease occurs or has an impact at an anatomic site.                                                    \\
disease has normal cell origin                & A disease originates from a type of normal cell.                                                          \\
disease has normal tissue origin              & A disease originates from a type of normal tissue.                                                        \\
disease mapped to gene                        & A gene is associated with a specific disease.                                                             \\
disease may have associated disease           & A disease may be associated with another disease.                                                         \\
disease may have finding                      & A possible clinical finding or symptom is observed in a disease.                                          \\
disease may have molecular abnormality        & A potential molecular abnormalities may be present in a disease.                                          \\
gene encodes gene product                     & A particular gene encodes a specific gene product, such as protein.                                       \\
gene product has associated anatomy           & A gene product is associated to an anatomical structure.                                                  \\
gene product has biochemical function         & A gene product is associated to a biochemical function.                                                   \\
gene product plays role in biological process & A gene product plays a role in a biological process.                                                      \\
has physiologic effect                        & A substance or process has a physiological effect on the body.                                            \\
may prevent                                   & A substance  may prevent a disease.                                                                       \\
may treat                                     & A substance  may treat a disease.                                                                         \\
occurs after                                  & A event or condition occurs after another.                                                                \\ \hline
\end{tabular}
\caption{Relation types in the MedLAMA dataset that involve in our study.}
\label{tab:relmedlama}
\end{table}
\begin{table}[t]
\centering
\begin{tabular}{lm{10cm}}
\hline
Relation Type       & Description                                                                                                 \\ \hline
Symptoms            & Physical or mental feature that indicates the presence of the disease.                                      \\
Affected sites      & Specific parts of the body that are impacted or harmed by the disease.                                      \\
Therapeutic Drugs   & Pharmaceutical substances prescribed to manage, alleviate, or cure the symptoms and effects of the disease. \\
Surgical Procedures & Medical procedures that treat the disease, involving the cutting, repairing, or removal of body parts.      \\ \hline
\end{tabular}
\caption{Relation types involved in the DiseK dataset.}
\label{tab:reldisek}
\end{table}
\begin{table}[t]
\centering
\setlength{\tabcolsep}{2mm}{
\begin{tabular}{lcc}
\hline
Relation Type     & \# Positive & \# Negative  \\ \hline
\#Symptoms             & 987 & 987 \\
\#Affected Sites      & 745 & 745  \\
\#Therapeutic Drugs            & 836 & 836  \\
\#Surgical Procedures & 599 & 599  \\ \hline
\end{tabular}}
\caption{Statistics of the sampled DiseK dataset. \# Positive: the number of positive triplets extracted from DiseK. \# Negative: the number of negative triplets sampled from DiseK. }
\label{tab:statistics}
\end{table}

\begin{table}[h]
    \centering
    \begin{tabular}{lcc}
    \hline
    Dataset      & MedLAMA & DiseK \\ \hline
    Type         & Biomedical & Clinical \\
    \# Rel Types & 17         & 4        \\
    \# Triplets  & 34,000    & 6,348   \\ \hline
    \end{tabular}
    \caption{Statistics of the sampled datasets.}
    \label{tab:datasets}
\end{table}

\begin{table}[t]
\centering
\begin{tabular}{cc}
\hline
Categories & Keywords\\
\hline
True    &   True, Entailed, Correct, Yes\\
False   &   False, Contradicted, Wrong, No\\
\hline
\end{tabular}
\caption{The keywords we utilize to extract answers from LLMs' responses.}
\label{tab: keywords}
\end{table}

\section{Effect of Rephrasing LLM Selection}\label{apdix:reph}
We have also leveraged Phi-3-medium-4k-instruct \citep{abdin2024phi} as the rephrasing model in our prototype-based sample generation module to study the effect of rephrasing LLM selection on the evaluation results. The evaluation results in Table \ref{tab:reph_effect} show that LLMs generally achieve similar performance on datasets generated based on different rephrasing LLMs, indicating that the effect of rephrasing LLM selection is minimal to the final evaluation results.
\begin{table}[t]
\centering
\begin{tabular}{lcccc}
\hline
\multirow{2}{*}{Model} & \multicolumn{2}{c}{MedLAMA} & \multicolumn{2}{c}{DiseK} \\
                       & Llama-3 Reph  & Phi-3 Reph  & Llama-3 Reph & Phi-3 Reph \\ \hline
Llama2-7B              & +3.1          & +2.9        & +2.8         & +2.3       \\
Vicuna-7B              & +7.5          & +6.3        & +3.9         & +3.5       \\
Vicuna-13B             & +10.7         & +9.8        & +5.7         & +5.3       \\
Gemma-7B               & +9.4         & +9.1        & +5.0         & +5.3       \\
Llama3-8B              & +16.6         & +16.4       & +9.3         & +9.3       \\
Llama2-70B             & +13.8         & +13.3       & +9.0         & +7.0       \\
Clinicalcamel-70B      & +21.9         & +22.4       & +16.1        & +14.6      \\
Meditron-70B           & +14,7         & +15.8       & +10.2         & +8.2       \\
Med42-70B              & +20.0         & +20.4       & +14.8        & +13.9      \\
Llama3-70B             & +26.9         & +27.4       & +20.9        & +19.3      \\
GPT-3.5-turbo          & +16.2         & +17.9       & +10.3         & +8.2       \\ 
GPT-4                  & +31.7         & +32.3       & +26.7         & +26.1\\\hline
\end{tabular}
\caption{Performance of LLMs on datasets generated by PretexEval using different rephrasing LLMs.}
\label{tab:reph_effect}
\end{table}
\section{Details of Method Setting}\label{apdix:framework}
\paragraph{Details of Predicate Equivalent Transformation}
\begin{figure}[h]
    \centering
    \includegraphics[width=0.49\textwidth]{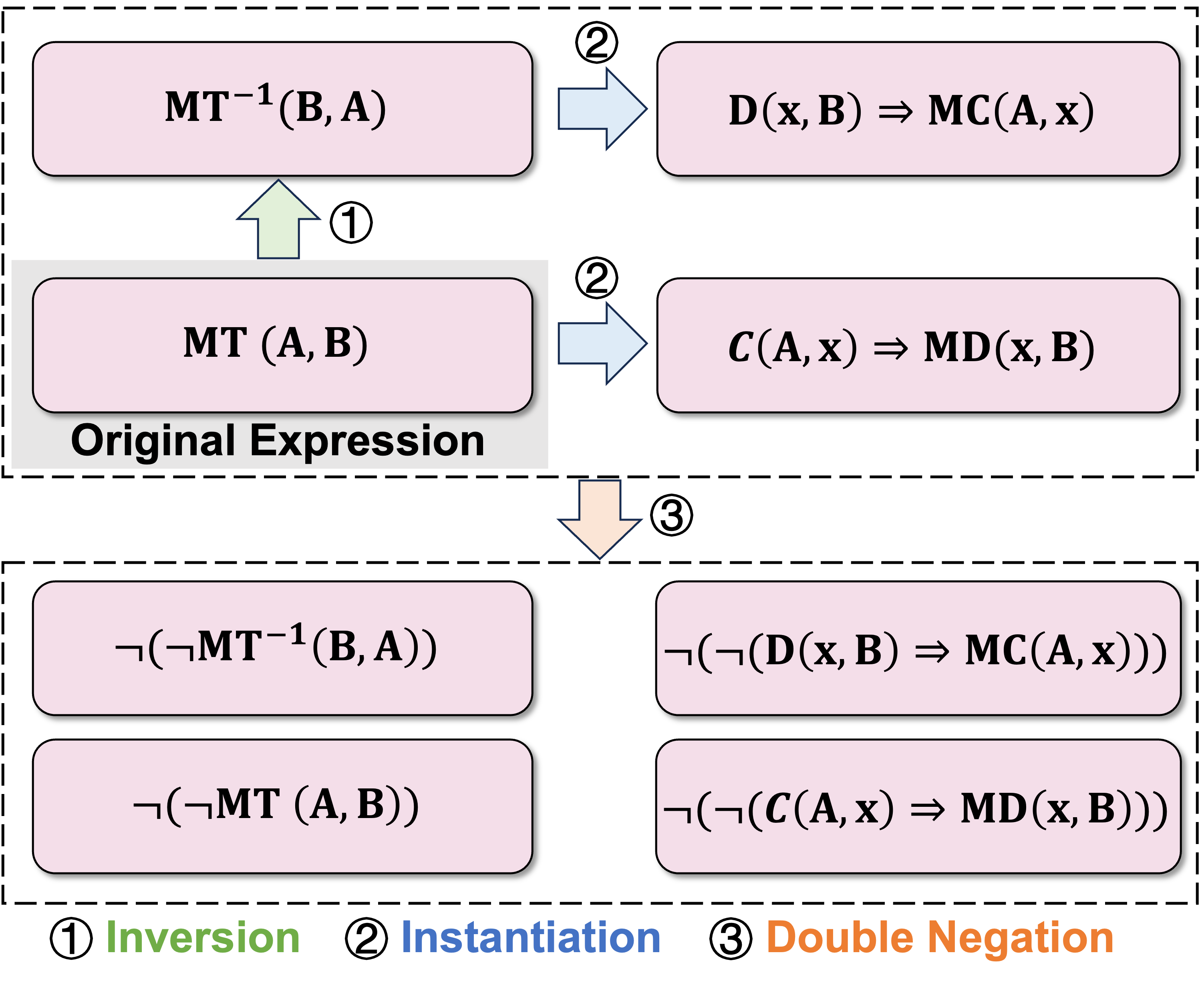}
    \caption{An example of the predicate equivalent transformation procedure implemented in this study.}
    \label{fig:procedure}
\end{figure}
An example of the predicate equivalent transformation procedure applied in this study is illustrated in Figure \ref{fig:procedure}. First, the inversion operation is applied to the original expression to create a new expression. Subsequently, these two expressions are instantiated into two additional expressions. Finally, double negation is used to generate four more expressions. 
\paragraph{Details of Prototypes-based Generation}
As introduced before, we designed a prototype-based sample generation strategy to ensure the reliability of the generated samples and crafted a prototype for each combination of relation type and predicate transformation type by discussing with clinicians. We list all the crafted prototypes in Table \ref{tab:prototype_impmedlama}, \ref{tab:prototype_impmedlama_1}, and \ref{tab: stmt} for reproducing our experiments. 

For LLM rephrasing, we prompt the Llama3-70B-Instruct model with the following instruction: ``\textit{Please paraphrase the following statement to present the same concept in a different way. DO NOT change the basic sentence structure. Directly output the paraphrased statement without other text. Statement: [prototype]}". In our experiments, we found that statements rephrased using this method effectively preserve the original meaning of the prototypes.
\begin{table*}[t]
\centering
\resizebox{\textwidth}{!}{
\begin{tabular}{m{2cm}m{2cm}m{3cm}m{4cm}m{4cm}}
\hline
\multicolumn{1}{c}{\multirow{2}{*}{Relation Type}} & \multicolumn{4}{c}{Predicate Transformation Type}                                                                                                                                                                                                                                                                                                                                                             \\\cline{2-5}
\multicolumn{1}{c}{}                               & None                                                            & Inv                                                                            & Ins                                                                                                                & Inv+Ins                                                                                                                 \\ \hline
associated morphology of                           & {[}X{]} is the associated morphology of {[}Y{]} .                        & {[}Y{]} is often accompanied by the morphology of {[}X{]}.                     & If a patient exhibits a morphological change of {[}X{]}, then he/she may suffer from {[}Y{]}.                      & If a patient suffers from {[}Y{]}, then he/she is exhibiting a morphological change of {[}X{]}.                         \\\hline
disease has abnormal cell                          & {[}X{]} has the abnormal cell {[}Y{]} .                                  & The abnormal cell type {[}Y{]} is detected within {[}X{]}.                     & If a patient suffers from {[}X{]}, then he/she has the abnormal cell {[}Y{]}.                                      & If a patient has the abnormal cell {[}Y{]}, then he/she may suffer from {[}X{]}.                                        \\\hline
disease has associated anatomic site               & The disease {[}X{]} can stem from the associated anatomic site {[}Y{]} . & Anatomical site {[}Y{]} is associated with the development of disease {[}X{]}. & If a patient suffers from {[}X{]}, then he/she has lesions in {[}Y{]}.                                             & If a patient has lesions in {[}Y{]}, then he/she may suffer from {[}X{]}.                                               \\\hline
disease has normal cell origin                     & The disease {[}X{]} stems from the normal cell {[}Y{]} .                 & Normal cell {[}Y{]} is associaated with the development of disease {[}X{]}.    & If a patient suffers from {[}X{]}, then he/she has lesions in {[}Y{]}.                                             & If a patient has lesions in {[}Y{]}, then he/she may suffer from {[}X{]}.                                               \\\hline
disease has normal tissue origin                   & The disease {[}X{]} stems from the normal tissue {[}Y{]} .               & Normal tissue {[}Y{]} is associated with the development of disease {[}X{]}.   & If a patient suffers from {[}X{]}, then he/she has lesions in {[}Y{]}.                                             & If a patient has lesions in {[}Y{]}, then he/she may suffer from {[}X{]}.                                               \\\hline
disease mapped to gene                             & The disease {[}X{]} is mapped to gene {[}Y{]} .                          & Gene {[}Y{]} is associated with the disease {[}X{]}.                           & If a patient suffers from {[}X{]}, then he/she has lesions in {[}Y{]}.                                             & If a patient has lesions in {[}Y{]}, then he/she may suffer from {[}X{]}.                                               \\\hline
disease may have associated disease                & The disease {[}X{]} might have the associated disease {[}Y{]} .          & The disease {[}Y{]} might have the associated disease {[}X{]} .                & If a patient suffers from {[}X{]}, then the likelihood of he/she suffering from {[}Y{]} is higher.                 & If a patient suffers from {[}Y{]}, then the likelihood of he/she suffering from {[}X{]} is higher.                      \\\hline
disease may have finding                           & {[}X{]} may have {[}Y{]} .                                               & {[}Y{]} may be associate with {[}X{]}                                          & If a patient suffers from {[}X{]}, then he/she has {[}Y{]}.                                                        & If a patient has {[}Y{]}, then he/she may suffer from {[}X{]}.                                                          \\\hline
disease may have molecular abnormality             & The disease {[}X{]} may have molecular abnormality {[}Y{]} .             & Molecular abnormality {[}Y{]} may be associated with the disease {[}X{]}.      & If a patient suffers from {[}X{]}, then he/she may has molecular abnormality {[}Y{]}.                              & If a patient has molecular abnormality {[}Y{]}, then he/she may suffer from {[}X{]}.                                    \\\hline
gene encodes gene product                          & The gene {[}X{]} encodes gene product {[}Y{]} .                          & The gene product {[}Y{]} is encoded by the gene {[}X{]}.                       & If the expression level of {[}X{]} decreases, it may lead to a reduction in the production or activity of {[}Y{]}. & If the production or activity of {[}Y{]} decreases, it may caused by the reduction in the expression level of  {[}X{]}. \\\hline
gene product has associated anatomy                & The gene product {[}X{]} has the associated anatomy {[}Y{]} .            & The anatomy {[}Y{]} is associated with the gene product {[}X{]}.               & The gene product {[}X{]} plays a role in anatomy {[}Y{]}.                                                          & Anatomy {[}Y{]} is where {[}X{]} functions.                                                                             \\\hline
gene product has biochemical function              & {[}X{]} has biochemical function {[}Y{]} .                               & {[}Y{]} is a biochemical function of {[}X{]}.                                  & If the production of {[}X{]} decreases, the functionality of {[}Y{]} may decrease.                                 & If the functionality of {[}Y{]} decreases, it may caused by the reduction in the production of {[}X{]}.                 \\\hline
gene product plays role in biological process      & The gene product {[}X{]} plays a role in biological process {[}Y{]} .    & Biological process {[}Y{]} is associated with the gene product {[}X{]}         & If the production of {[}X{]} decreases, the process of {[}Y{]} may be influenced.                                  & If {[}Y{]} is affected, it may caused by the reduction in the production of {[}X{]}.                                    \\\hline
has physiologic effect                             & {[}X{]} has physiologic effect of {[}Y{]} .                              & {[}Y{]} can be caused by {[}X{]}.                                              & If a patient takes {[}X{]}, he/she may have physiologic effect of {[}Y{]} .                                        & If a patient has physiologic effect of {[}Y{]}, he/she may have taken {[}X{]}.                                          \\\hline
may prevent                                        & {[}X{]} may be able to prevent {[}Y{]} .                                 & {[}Y{]} may be prevented by {[}X{]}                                            & If a patient takes {[}X{]}, he/she can prevent {[}Y{]}.                                                            & If a patient wishes to prevent {[}Y{]}, he/she should take {[}X{]}.                                                     \\\hline
may treat                                          & {[}X{]} might treat {[}Y{]} .                                            & {[}Y{]} may be treated by {[}X{]}                                              & If a patient takes {[}X{]}, he/she can treat {[}Y{]}.                                                              & If a patient suffers from {[}Y{]}, he/she should take {[}X{]}.                                                          \\\hline
occurs after                                       & {[}X{]} occurs after {[}Y{]} .                                           & {[}Y{]} may occur before {[}X{]}.                                              & If a patient occurs {[}X{]}, he/she may occur {[}Y{]} before.                                                      & If a patient occurs {[}Y{]}, he/she may occur {[}X{]} afterwards.                                                       \\ \hline
\end{tabular}}
\caption{Prototypes crafted for the MedLAMA dataset (1/2). Inv: inversion; Ins: instantiation.}
\label{tab:prototype_impmedlama}
\end{table*}

\begin{table*}[t]
\centering
\small
\resizebox{\textwidth}{!}{
\begin{tabular}{m{2cm}m{2cm}m{3cm}m{4cm}m{4cm}}
\hline
\multicolumn{1}{c}{\multirow{2}{*}{Relation Type}} & \multicolumn{4}{c}{Predicate Transformation Type}                                                                                                                                                                                                                                                                                                                                                             \\\cline{2-5}
\multicolumn{1}{c}{}                               & DN                                                                            & Inv+DN                                                                             & Ins+DN                                                                                                 & Inv+Ins+DN                                                                                                                  \\ \hline
associated morphology of                           & {[}X{]} is not the associated morphology of {[}Y{]}.                          & {[}Y{]} is not accompanied by the morphology of {[}X{]}.                           & A patient that exhibits a morphological change of {[}X{]} does not suffer from {[}Y{]}.                & A patient that suffers from {[}Y{]} does not exhibit a morphological change of {[}X{]}.                                 \\\hline
disease has abnormal cell                          & {[}X{]} does not has the abnormal cell {[}Y{]}.                               & The abnormal cell type {[}Y{]} is not detected within {[}X{]}.                     & A patient that suffers from {[}X{]} does not have the abnormal cell {[}Y{]}.                           & A patient that has the abnormal cell {[}Y{]} does not suffer from {[}X{]}.                                              \\\hline
disease has associated anatomic site               & The disease {[}X{]} is not stem from the associated anatomic site {[}Y{]}.    & Anatomical site {[}Y{]} is not associated with the development of disease {[}X{]}. & A patient that suffers from {[}X{]} does not have lesions in {[}Y{]}.                                  & A patient that has lesions in {[}Y{]} does not suffer from {[}X{]}.                                                     \\\hline
disease has normal cell origin                     & The disease {[}X{]} does not stem from the normal cell {[}Y{]}.               & Normal cell {[}Y{]} is not associaated with the development of disease {[}X{]}.    & A patient that suffers from {[}X{]} does not have lesions in {[}Y{]}.                                  & A patient that has lesions in {[}Y{]} does not suffer from {[}X{]}.                                                     \\\hline
disease has normal tissue origin                   & The disease {[}X{]} is not stem from the normal tissue {[}Y{]}.               & Normal tissue {[}Y{]} is not associated with the development of disease {[}X{]}.   & A patient that suffers from {[}X{]} does not have lesions in {[}Y{]}.                                  & A patient that has lesions in {[}Y{]} does not suffer from {[}X{]}.                                                     \\\hline
disease mapped to gene                             & The disease {[}X{]} is not mapped to the gene {[}Y{]}.                        & Gene {[}Y{]} is not associated with the disease {[}X{]}.                           & A patient that suffers from {[}X{]} does not have lesions in {[}Y{]}.                                  & A patient that has lesions in {[}Y{]} does not suffer from {[}X{]}.                                                     \\\hline
disease may have associated disease                & The disease {[}X{]} is not associated with disease {[}Y{]} .                  & The disease {[}Y{]} is not associated with disease {[}X{]} .                       & If a patient suffers from {[}X{]}, then the likelihood of he/she suffering from {[}Y{]} is not higher. & If a patient suffers from {[}Y{]}, then the likelihood of he/she suffering from {[}X{]} is not higher.                  \\\hline
disease may have finding                           & {[}X{]} does not have {[}Y{]} .                                               & {[}Y{]} is not associated with {[}X{]}                                             & A patient that suffers from {[}X{]} does not have {[}Y{]}.                                             & A patient that has {[}Y{]} does not suffer from {[}X{]}.                                                                \\\hline
disease may have molecular abnormality             & The disease {[}X{]} does not have molecular abnormality {[}Y{]} .             & Molecular abnormality {[}Y{]} is not associated with the disease {[}X{]}.          & A patient that suffers from {[}X{]} does not have molecular abnormality {[}Y{]}.                       & A patient that has molecular abnormality {[}Y{]} does not suffer from {[}X{]}.                                          \\\hline
gene encodes gene product                          & The gene {[}X{]} does not encode gene product {[}Y{]} .                       & The gene product {[}Y{]} is not encoded by the gene {[}X{]}                        & A decrease in the expression level of {[}X{]} does not affect the production and activity of {[}Y{]}.  & A decrease in the production or activity of {[}Y{]} is not caused by the reduction in the expression level of  {[}X{]}. \\\hline
gene product has associated anatomy                & The gene product {[}X{]} does not have the associated anatomy {[}Y{]} .       & The anatomy {[}Y{]} is not associated with the gene product {[}X{]}.               & The gene product {[}X{]} does not play a role in anatomy {[}Y{]}.                                      & Anatomy {[}Y{]} is not where {[}X{]} functions.                                                                         \\\hline
gene product has biochemical function              & {[}X{]} does not have biochemical function {[}Y{]} .                          & {[}Y{]} is not a biochemical function of {[}X{]}.                                  & A decrease in the production of {[}X{]} does not affect the functionality of {[}Y{]}.                  & A decrease in the functionality of {[}Y{]} is not caused by the reduction in the production of {[}X{]}.                 \\\hline
gene product plays role in biological process      & The gene product {[}X{]} does not play a role in biological process {[}Y{]} . & Biological process {[}Y{]} is not associated with the gene product {[}X{]}         & A decrease in the production of {[}X{]} does not affect the process of {[}Y{]}.                        & A change of {[}Y{]} is not caused by the reduction in the production of {[}X{]}.                                        \\\hline
has physiologic effect                             & {[}X{]} does not have physiologic effect of {[}Y{]} .                         & {[}Y{]} cannot be caused by {[}X{]}.                                               & A patient that takes {[}X{]} does not have physiologic effect of {[}Y{]} .                             & A patient that has physiologic effect of {[}Y{]} has not taken {[}X{]}.                                                 \\\hline
may prevent                                        & {[}X{]} is not able to prevent {[}Y{]} .                                      & {[}Y{]} cannot be prevented by {[}X{]}                                             & Taking {[}X{]} have no effect on preventing {[}Y{]}.                                                   & A patient wishes to prevent {[}Y{]} has no need to take {[}X{]}.                                                        \\\hline
may treat                                          & {[}X{]} is not able to treat {[}Y{]} .                                        & {[}Y{]} cannot be treated by {[}X{]}                                               & Taking {[}X{]} have no effect on treating {[}Y{]}.                                                     & A patient that suffers from {[}Y{]}  has no need to take {[}X{]}.                                                       \\\hline
occurs after                                       & {[}X{]} does not occur after {[}Y{]} .                                        & {[}Y{]} cannot occur before {[}X{]}.                                               & A patient occurs {[}X{]} will not occur {[}Y{]} before.                                                & A patient occurs {[}Y{]} will not occur {[}X{]} afterwards.                       \\\hline
\end{tabular}}
\caption{Prototypes crafted for the MedLAMA dataset (2/2). Inv: inversion; Ins: instantiation; DN: double negation.}
\label{tab:prototype_impmedlama_1}
\end{table*}
\begin{table*}[t]
\centering
\resizebox{\textwidth}{!}{
\begin{tabular}{m{2cm}m{3cm}m{3cm}m{3cm}m{3cm}}
\hline
\multicolumn{1}{c}{\multirow{2}{*}{Implication Type}} & \multicolumn{4}{c}{Relation Type}                                                                                                                                                                                                                                                                                                     \\ \cline{2-5} 
\multicolumn{1}{c}{}                                          & Symptoms                                                                    & Affected Sites                                                                          & Therapeutic Drugs                                                             & Surgical Procedures                                                           \\ \hline
None                                                          & {[}Y{]} is a common symptom of {[}X{]}.                                     & {[}Y{]} is the affected site for {[}X{]}.                                                                                  & {[}Y{]} is a common medication for {[}X{]}. & {[}Y{]} is a common procedure for {[}X{]}.                                   \\ \hline
Inv                                                           & Common symptoms of {[}X{]} include {[}Y{]}.                                 & Affected sites for {[}X{]} include {[}Y{]}.                                              & Common medications for treating {[}X{]} include {[}Y{]}. & Common procedures for treating {[}X{]} include {[}Y{]}.                                            \\ \hline
Ins                                                           & If a patient has {[}X{]}, they are very likely to have symptoms of {[}Y{]}. & If a patient has {[}X{]}, their {[}Y{]} site is very likely to show lesions.            & If a patient has {[}X{]}, {[}Y{]} can be used to treat their condition.       & If a patient has {[}X{]}, {[}Y{]} can be used to treat their condition.       \\ \hline
Inv+Ins                                                       & If a patient has symptoms of {[}Y{]}, they are very likely to have {[}X{]}. & If a patient shows lesions in their {[}Y{]} site, they are very likely to have {[}X{]}. & If {[}Y{]} can be used to treat a patient's condition, they may have {[}X{]}. & If {[}Y{]} can be used to treat a patient's condition, they may have {[}X{]}. \\ \hline
DN                                                            & {[}Y{]} is not a common symptom of {[}X{]}.                                 & {[}Y{]} is not the affected site for {[}X{]}.                                         & {[}Y{]} is not a common medication for {[}X{]}.   & {[}Y{]} is not a common procedure for {[}X{]}.                                                              \\ \hline
Inv+DN                                                        & Common symptoms of {[}X{]} do not include {[}Y{]}.                          & Affected sites for {[}X{]} do not include {[}Y{]}.                                   & Common medications for treating {[}X{]} do not include {[}Y{]}.    & Common procedures for treating {[}X{]} do not include {[}Y{]}.                              \\ \hline
Ins+DN                                                        & Patients with {[}X{]} are unlikely to have symptoms of {[}Y{]}.             & Patients with {[}X{]} are unlikely to show lesions in their {[}Y{]} site.             & Patients with {[}X{]} do not commonly use {[}Y{]} for treatment.   & Patients with {[}X{]} do not commonly use {[}Y{]} for treatment.                       \\ \hline
Inv+DN                                                        & Patients with symptoms of {[}Y{]} are unlikely to have {[}X{]}.             & Patients showing lesions in their {[}Y{]} site are unlikely to have {[}X{]}.           & Patients who can be treated with {[}Y{]} are unlikely to have {[}X{]}.  & Patients who can be treated with {[}Y{]} are unlikely to have {[}X{]}.              \\ \hline
\end{tabular}}
\caption{Prototypes crafted for the DiseK dataset. Inv: inversion; Ins: instantiation; DN: double negation.}
\label{tab: stmt}
\end{table*}
\begin{figure}[t]
    \centering
    \includegraphics[width=0.8\textwidth]{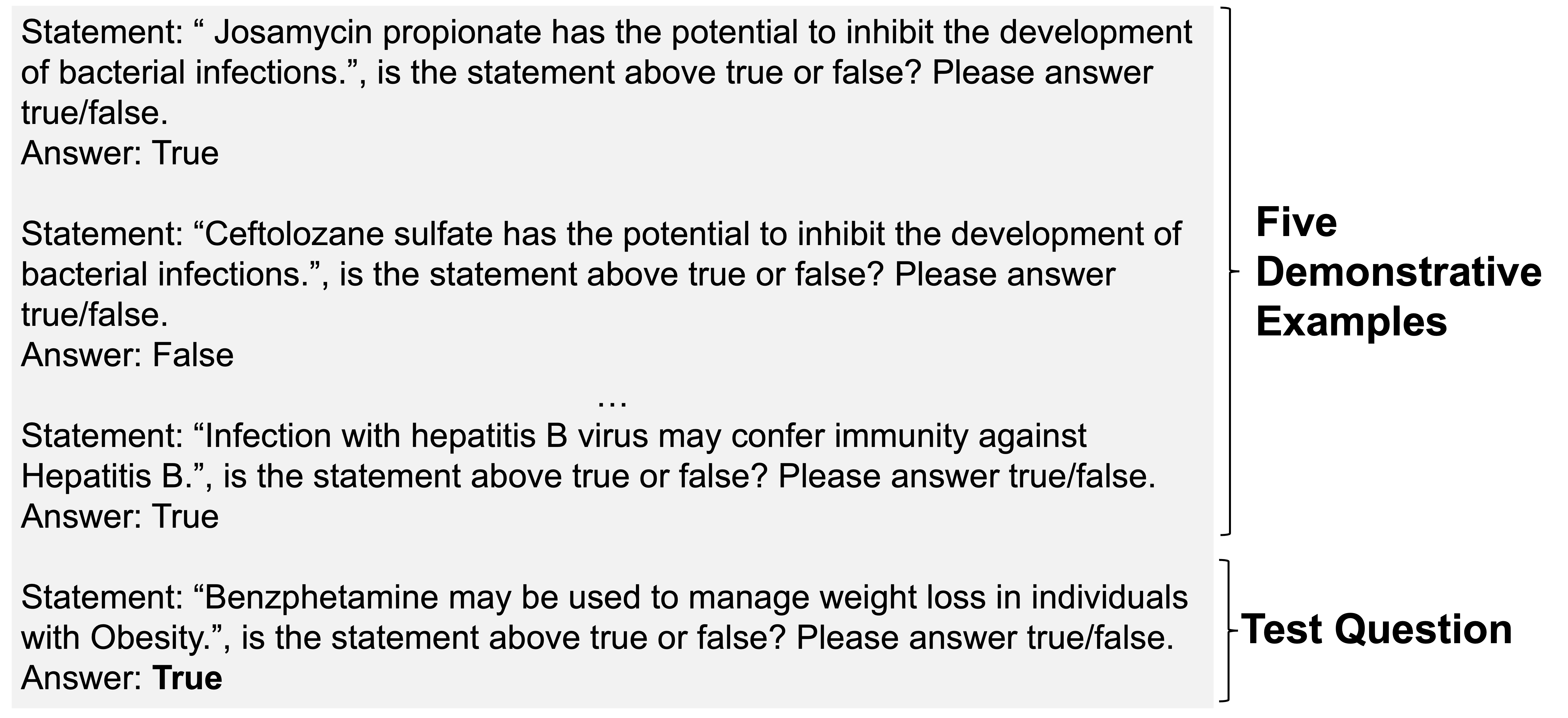}
    \caption{An example of the five-shot in-context learning process applied in our evaluation.}
    \label{fig: five-shot}
\end{figure}
\section{Details of Evaluation Setting}\label{apdix:eval}
In our implementation, we form test samples based on the following format: ``\textit{[Statement], is the statement above true or false? Please answer True or False.}" For the five-shot setting, we randomly select five question-answer pairs for each combination of relation and predicate transformation type to create demonstrative examples, as depicted in Figure \ref{fig: five-shot}. Complex prompting strategies such as chain-of-thought are not applied in our study, as the evaluation statements are crafted to be straightforward and easily understandable, allowing for verification without the need for complex logical reasoning. In the inference process, we use greedy search for most of LLMs. However, commercial LLMs like GPT-3.5-turbo do not support greedy search, and we use their default generation setting to make a relative fair comparison across LLMs. We extract the prediction from models' response based on keywords since the words/phrases used to express True and False are limited. We listed all of the keywords applied to recognize answers in Table \ref{tab: keywords}.
\section{Details of Baselines}\label{apdix:baseline}
We implement the LLMEval method by directly generating diverse statements using Llama3-70B-Instruct. Specifically, we prompt the LLM with the following instruction: ``\textit{Based on the given knowledge triplet, generate 8 statement to express the underlying knowledge in different ways. Output one statement per line. Directly output the statements without other text. Knowledge triplet: [triplet].}" To ensure the quality of generated samples, we use the greedy search for the decoding process. We find that Llama3-70B-Instruct can follow the instruction, generating samples in separated lines. 
\begin{table*}[t]
\centering
\resizebox{\textwidth}{!}{
\begin{tabular}{lcccccc}
\hline
\multicolumn{1}{c}{\multirow{2}{*}{Model}} & \multicolumn{3}{c}{MedLAMA}                                                                             & \multicolumn{3}{c}{DiseK}                                                                               \\
\multicolumn{1}{c}{}                       & PretexEval       & w/o PreEqTrans                                    & w/o LLM Rephrasing                                    & PretexEval       & w/o PreEqTrans                                    & w/o LLM Rephrasing                                    \\ \hline
Llama2-7B       &+3.1   &+$7.4\textsubscript{↑143.9\%}$ &+$1.9\textsubscript{↓36.6\%}$  &+2.8   &+$7.5\textsubscript{↑169.5\%}$ &+$2.6\textsubscript{↓7.4\%}$\\
Vicuna-7B       &+7.5   &+$22.1\textsubscript{↑193.0\%}$        &+$5.7\textsubscript{↓24.2\%}$  &+3.9   &+$9.5\textsubscript{↑142.5\%}$     &+$2.5\textsubscript{↓35.1\%}$\\
Vicuna-13B      &+10.7  &+$20.3\textsubscript{↑89.8\%}$ &+$11.0\textsubscript{↑3.3\%}$  &+5.7   &+$9.2\textsubscript{↑60.4\%}$  &+$5.9\textsubscript{↑2.7\%}$\\
Gemma-7B        &+9.4   &+$16.2\textsubscript{↑72.0\%}$ &+$12.8\textsubscript{↑35.6\%}$ &+5.0   &+$7.2\textsubscript{↑44.6\%}$  &+$6.9\textsubscript{↑39.9\%}$\\
Llama3-8B       &+16.6  &+$24.1\textsubscript{↑45.5\%}$ &+$18.5\textsubscript{↑11.9\%}$ &+9.3   &+$18.9\textsubscript{↑104.3\%}$   &+$10.2\textsubscript{↑9.7\%}$\\
Llama2-70B      &+13.8  &+$28.2\textsubscript{↑104.7\%}$        &+$14.6\textsubscript{↑5.8\%}$  &+9.0   &+$18.4\textsubscript{↑103.5\%}$    &+$7.8\textsubscript{↓14.1\%}$\\
ClinicalCamel-70B       &+21.9  &+$30.6\textsubscript{↑40.2\%}$ &+$22.8\textsubscript{↑4.5\%}$  &+16.1  &+$23.1\textsubscript{↑44.0\%}$     &+$18.0\textsubscript{↑12.0\%}$\\
Meditron-70B    &+14.7  &+$25.7\textsubscript{↑74.6\%}$ &+$15.8\textsubscript{↑7.1\%}$  &+10.2  &+$18.1\textsubscript{↑77.0\%}$ &+$11.5\textsubscript{↑12.7\%}$\\
Med42-70B       &+20.0  &+$28.2\textsubscript{↑40.7\%}$ &+$20.4\textsubscript{↑1.9\%}$  &+14.8  &+$20.4\textsubscript{↑38.3\%}$ &+$17.9\textsubscript{↑21.0\%}$\\
GPT-3.5-turbo   &+16.7  &+$29.4\textsubscript{↑76.0\%}$ &+$17.9\textsubscript{↑7.3\%}$  &+10.3  &+$17.1\textsubscript{↑66.5\%}$ &+$11.8\textsubscript{↑15.6\%}$\\
Llama3-70B      &+26.9  &+$33.0\textsubscript{↑22.8\%}$ &+$30.4\textsubscript{↑13.3\%}$ &+20.9  &+$27.8\textsubscript{↑33.3\%}$ &+$24.0\textsubscript{↑15.1\%}$\\
GPT-4o  &+31.7  &+$36.0\textsubscript{↑13.7\%}$ &+$33.8\textsubscript{↑6.7\%}$  &+26.7  &+$29.2\textsubscript{↑9.5\%}$      &+$30.4\textsubscript{↑13.8\%}$\\\hline
\end{tabular}}
\caption{Ablation results of all evaluated LLMs for key components of the proposed PretexEval framework. PreEqTrans: Predicate Equivalence Transformation; LLM Rephrasing: Prototype-based Sample Generation.}
\label{tab:ablation_all}
\end{table*}
\begin{table*}[t]
\centering
\resizebox{\textwidth}{!}{
\begin{tabular}{lcccccccc}
\hline
\multicolumn{1}{c}{\multirow{2}{*}{Model}} & \multicolumn{4}{c}{MedLAMA}                                                                             & \multicolumn{4}{c}{DiseK}                                                                               \\
\multicolumn{1}{c}{}                       & None       & +Inv & +DN+Inv & +All  & Origin       & +Inv & +DN+Inv & +All                                    \\ \hline
Llama2-7B       &+7.4   &+$7.1\textsubscript{↓5.0\%}$   &+$3.6\textsubscript{↓51.8\%}$  &+$3.1\textsubscript{↓59.0\%}$      &+7.5   &+$7.9\textsubscript{↑4.8\%}$   &+$3.9\textsubscript{↓48.5\%}$  &+$2.8\textsubscript{↓62.9\%}$\\
Vicuna-7B       &+22.1  &+$21.8\textsubscript{↓1.0\%}$  &+$8.2\textsubscript{↓62.9\%}$  &+$7.5\textsubscript{↓65.9\%}$      &+9.5   &+$11.4\textsubscript{↑20.3\%}$ &+$4.7\textsubscript{↓50.2\%}$  &+$3.9\textsubscript{↓58.8\%}$\\
Vicuna-13B      &+20.3  &+$19.9\textsubscript{↓1.9\%}$  &+$11.6\textsubscript{↓43.0\%}$ &+$10.7\textsubscript{↓47.3\%}$     &+9.2   &+$10.0\textsubscript{↑9.0\%}$  &+$5.8\textsubscript{↓37.3\%}$  &+$5.7\textsubscript{↓37.7\%}$\\
Gemma-7B        &+16.2  &+$15.9\textsubscript{↓2.1\%}$  &+$10.8\textsubscript{↓33.5\%}$ &+$9.4\textsubscript{↓41.9\%}$      &+7.2   &+$10.4\textsubscript{↑44.6\%}$ &+$5.2\textsubscript{↓27.6\%}$  &+$5.0\textsubscript{↓30.9\%}$\\
Llama3-8B       &+24.1  &+$23.3\textsubscript{↓3.2\%}$  &+$18.5\textsubscript{↓23.2\%}$ &+$16.6\textsubscript{↓31.3\%}$     &+18.9  &+$18.6\textsubscript{↓1.8\%}$  &+$10.1\textsubscript{↓46.7\%}$ &+$9.3\textsubscript{↓51.1\%}$\\
Llama2-70B      &+28.2  &+$27.4\textsubscript{↓2.9\%}$  &+$15.8\textsubscript{↓43.8\%}$ &+$13.8\textsubscript{↓51.2\%}$     &+18.4  &+$18.8\textsubscript{↑2.1\%}$  &+$9.7\textsubscript{↓47.1\%}$  &+$9.0\textsubscript{↓50.9\%}$\\
ClinicalCamel-70B       &+30.6  &+$30.3\textsubscript{↓1.1\%}$  &+$23.2\textsubscript{↓24.2\%}$ &+$21.9\textsubscript{↓28.7\%}$     &+23.1  &+$22.4\textsubscript{↓3.1\%}$  &+$17.9\textsubscript{↓22.5\%}$ &+$16.1\textsubscript{↓30.5\%}$\\
Meditron-70B    &+25.7  &+$25.4\textsubscript{↓1.2\%}$  &+$15.8\textsubscript{↓38.6\%}$ &+$14.7\textsubscript{↓42.7\%}$     &+18.1  &+$19.5\textsubscript{↑7.8\%}$  &+$11.1\textsubscript{↓38.9\%}$ &+$10.2\textsubscript{↓43.5\%}$\\
Med42-70B       &+28.2  &+$27.9\textsubscript{↓1.1\%}$  &+$21.9\textsubscript{↓22.3\%}$ &+$20.0\textsubscript{↓28.9\%}$     &+20.4  &+$20.2\textsubscript{↓1.1\%}$  &+$15.7\textsubscript{↓23.1\%}$ &+$14.8\textsubscript{↓27.7\%}$\\
GPT-3.5-turbo   &+29.4  &+$27.6\textsubscript{↓6.3\%}$  &+$18.2\textsubscript{↓38.0\%}$ &+$16.7\textsubscript{↓43.2\%}$     &+17.1  &+$18.1\textsubscript{↑6.1\%}$  &+$9.6\textsubscript{↓43.8\%}$  &+$10.3\textsubscript{↓39.9\%}$\\
Llama3-70B      &+33.0  &+$31.8\textsubscript{↓3.6\%}$  &+$28.6\textsubscript{↓13.2\%}$ &+$26.9\textsubscript{↓18.6\%}$     &+27.8  &+$27.5\textsubscript{↓1.4\%}$  &+$22.3\textsubscript{↓19.8\%}$ &+$20.9\textsubscript{↓25.0\%}$\\
GPT-4o  &+36.0  &+$34.2\textsubscript{↓4.9\%}$  &+$33.6\textsubscript{↓6.8\%}$  &+$31.7\textsubscript{↓12.1\%}$     &+29.2  &+$29.8\textsubscript{↑1.7\%}$  &+$26.8\textsubscript{↓8.3\%}$  &+$26.7\textsubscript{↓8.7\%}$\\ \hline
\end{tabular}}
\caption{Ablation results of all evaluated LLMs for types of predicate transformation in the proposed framework.}
\label{tab:ablation_all_types}
\end{table*}

\section{Performance of LLMs Across Knowledge Types}\label{apdix:types}
We also analysis the fine-grained performance of LLMs across different types of clinical knowledge stored in DiseK. The analysis results are presented in Table \ref{tab:disek_detail_perf}. Based on this analysis, we can draw the following conclusions: (1) GPT-4o outperforms the rest of LLMs on 3 out of 4 types of clinical knowledge, exhibiting more comprehensive mastery of medical knowledge than other LLMs; (2) With the same model parameter scale, Llama3-70B achieved the best performance across all four relation types, possibly due to its significantly large training data volume (7.5 times that of Llama2-70B); (3) The three medical-specific 70B models (ClinicalCamel, Meditron, Med42) are all developed based on Llama2-70B through finetuning on medical corpora, and they show a notable improvement in medical knowledge mastery compared to Llama2-70B. In particular, ClinicalCamel-70B enhanced accuracy on “Affected Sites” from 72.2 with Llama2-70B to 84.1, while Med42-70B improved performance on “Surgical Procedures,” raising it from 62.6 with Llama2-70B to 71.7. 
\begin{table}[t]
\centering
\begin{tabular}{lcccc}
\hline
Model             & Symptoms & Affected Sites & Therapeutic Drugs & Surgical Procedures \\ \hline
Llama2-7B         & +1.9                         & +10.7                              & +-0.4                                 & +-1.1                                   \\
Vicuna-7B         & +0.2                         & +8.1                               & +1.3                                  & +8.6                                    \\
Vicuna-13B        & +3.3                         & +14.2                              & +1.1                                  & +5.9                                    \\
Gemma-7B          & +3.2                         & +11.2                              & +0.7                                  & +6.1                                    \\
Llama3-8B         & +6.2                         & +17.4                              & +3.8                                  & +11.8                                   \\ \hline
Llama2-70B        & +4.6                         & +22.2                              & +0.1                                  & +12.6                                   \\
ClinicalCamel-70B & +9.5                         & +34.1                              & +5.1                                  & +19.6                                   \\
Meditron-70B      & +6.9                         & +18.3                              & +3.0                                  & +15.9                                   \\
Med42-70B         & +7.2                         & +29.9                              & +5.4                                  & +21.7                                   \\
Llama3-70B        & +15.1                        & +37.7                              & +11.6                                 & \textbf{+22.4}                          \\
GPT-3.5-turbo     & +5.5                         & +19.4                              & +5.6                                  & +13.1                                   \\
GPT-4o            & \textbf{+23.4}               & \textbf{+41.5}                     & \textbf{+20.9}                        & +19.0                                   \\ \hline
Average           & +7.1                         & +21.2                              & +4.8                                  & +12.4                                   \\ \hline
\end{tabular}
\caption{Performance of LLMs on the four types of disease-related knowledge contained in the DiseK knowledge base.}
\label{tab:disek_detail_perf}
\end{table}
\section{Full Experimental Results}\label{apdix:experiments}
\begin{figure*}[t]
    \centering
    \includegraphics[width=\textwidth]{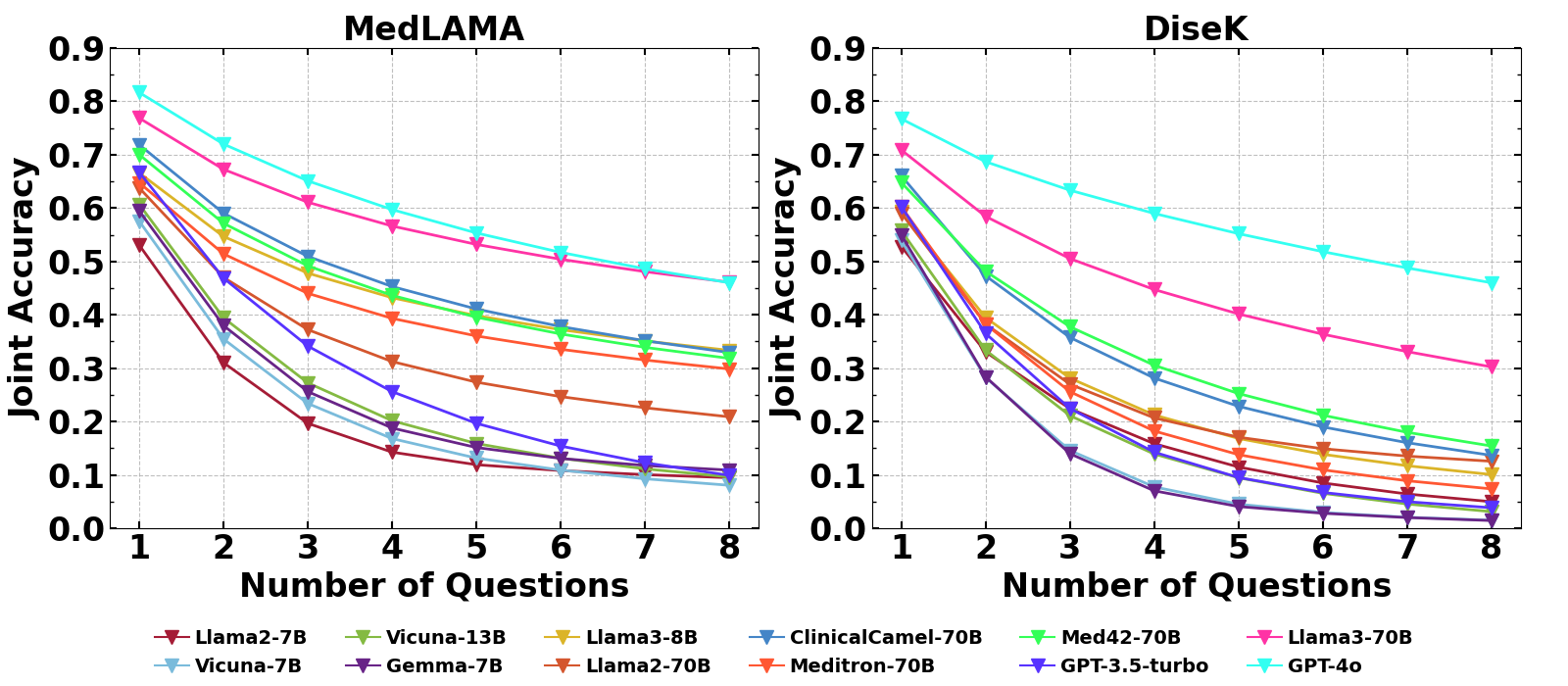}
    \caption{Performance (\textbf{joint accuracy}) of all LLMs evaluated by the proposed \textbf{PretexEval} framework.}
    \label{fig: PretexEval_perform}
\end{figure*}
\begin{figure*}[t]
    \centering
    \includegraphics[width=\textwidth]{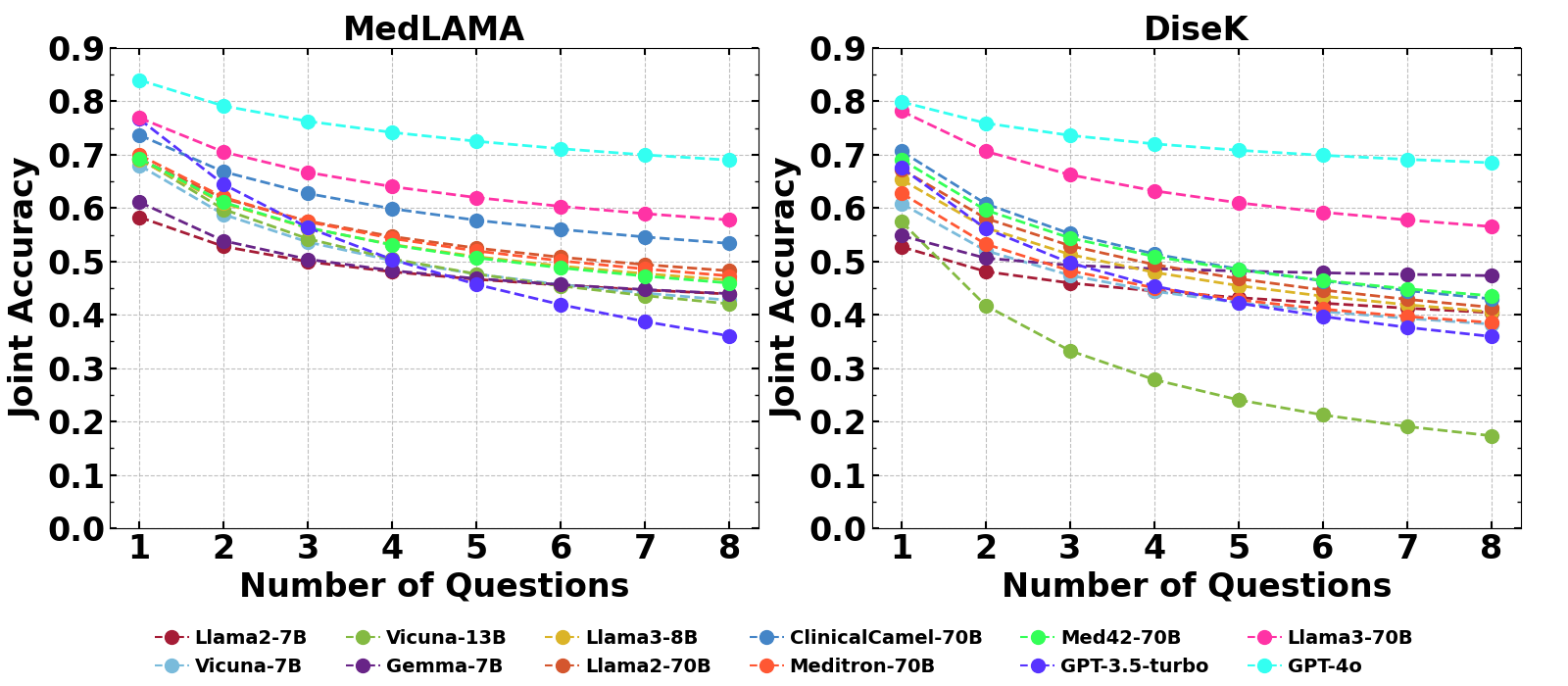}
    \caption{Performance (\textbf{joint accuracy}) of all LLMs evaluated by the \textbf{LLMEval} method.}
    \label{fig: llmeval_perform}
\end{figure*}
\subsection{Joint accuracy}\label{apdix:joint}
We illustrate the joint accuracy of all LLMs evaluated by PretexEval and LLMEval in Figure \ref{fig: PretexEval_perform} and \ref{fig: llmeval_perform}, respectively. The experimental results support our conclusions: the evaluated LLMs generally perform worse on datasets generated by PretexEval. Moreover, LLMs' performance decline faster when evaluated by PretexEval compared with evaluated by LLMEval, indicating that current LLMs lack consistency in understanding medical knowledge presented in various structures.
\subsection{Ablation Study}\label{apdix:ablation}
We also presents the ablation results of all evaluated LLMs regarding key components and predicate transformation types in Table \ref{tab:ablation_all} and \ref{tab:ablation_all_types}, respectively. These results are consistent with our findings in the paper, demonstrating the effectiveness of our framework.
\clearpage
\section{Details of Human Analysis on Generated Sample Quality}\label{apdix:quality}
\begin{figure}[h]
    \centering
    \includegraphics[width=0.8\linewidth]{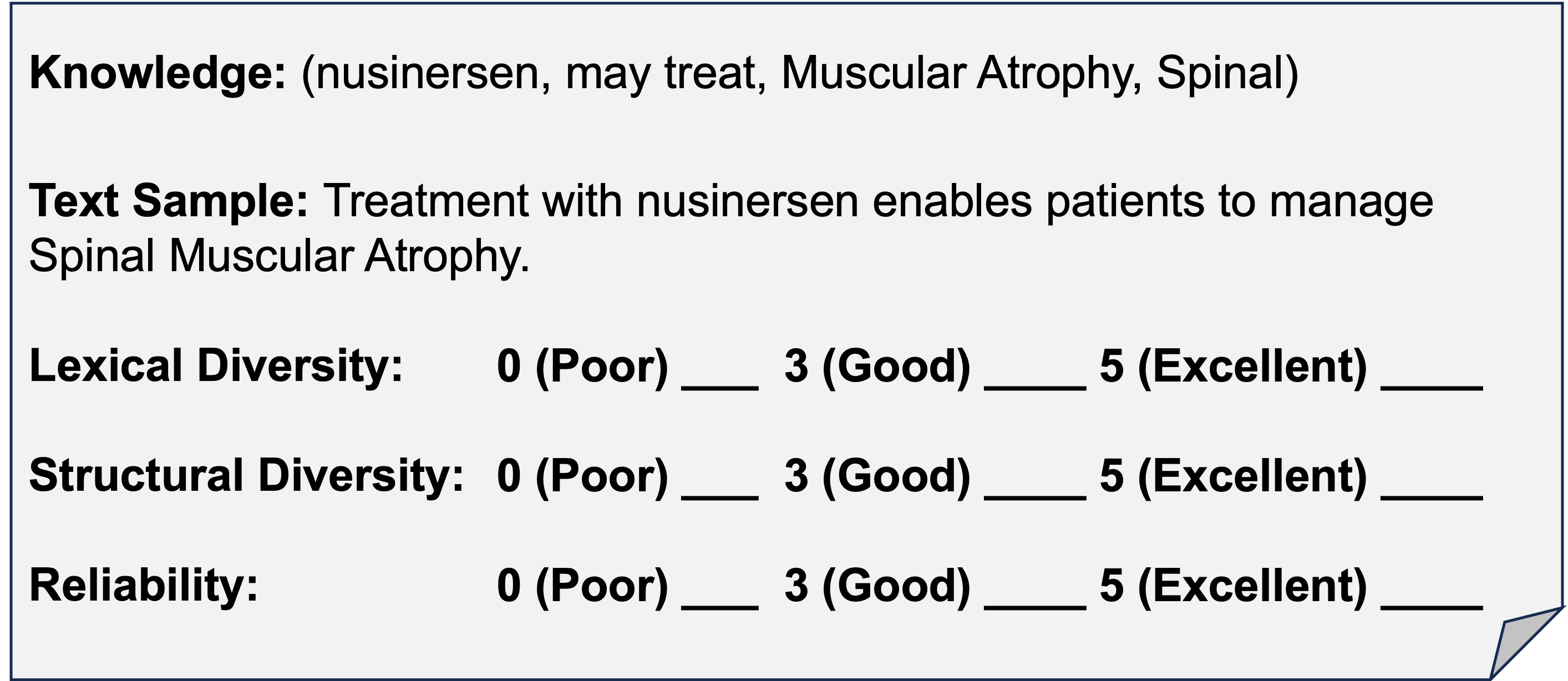}
    \caption{A grading sample presented to human doctors.}
    \label{fig:anno_sample}
\end{figure}

We conduct a human analysis on the quality of generated samples, regarding the reliability, lexical diversity, and structural diversity. As mentioned above, we randomly sampled 50 knowledge triplets from MedLAMA and collect the corresponding test samples generated by LLMEval and PretexEval before/after rephrasing. Then we have two experienced doctors to grade the test samples based on the following criteria:
\begin{enumerate}
    \item Reliability:
    \begin{itemize}
        \item 0 (Poor): Many inaccuracies are present, leading to significant misunderstandings or misinterpretations of the knowledge presented.
        \item 3 (Good): Most information is correct, but minor inaccuracies or ambiguities are present that do not affect the overall meaning.
        \item 5 (Excellent): All information presented is factually correct, with clear and precise explanations. No errors or ambiguities are detected.
    \end{itemize}
    \item Lexical Diversity:
    \begin{itemize}
        \item 0 (Poor): The text shows very limited vocabulary diversity compared to the original knowledge triplet.
        \item 3 (Good): There is a moderate variety of vocabulary compared to the original in the non-medical lexicon, while the medical terms remaining unchanged.
        \item 5 (Excellent): The text uses diverse vocabulary compared to the original knowledge triplet, including both medical terms and non-medical lexicon.
    \end{itemize}
    \item Structural Diversity:
    \begin{itemize}
        \item 0 (Poor): The sentence structure remains unchanged, fully replicating the original order of the knowledge triplet.
        \item 3 (Good): The sentence structure has been slightly adjusted, such as by changing word order or modifying certain phrase combinations, while the main grammatical structure remains unchanged.
        \item 5 (Excellent): The sentence structure has been thoroughly reconstructed, significantly altering the way information is presented, while conveying the same content with a completely new syntax and grammatical logic.
    \end{itemize}
\end{enumerate}
For each test sample, the doctors are presented with the original knowledge triplet for reference (see Figure \ref{fig:anno_sample}). We hide the source of each text sample to ensure the fairness of the evaluation. Finally, we average the scores of samples generated by the same method to derive the final scores.

\end{document}